\documentclass[12pt]{l4dc2021}

\usepackage{algorithm}
\usepackage{bm}
\usepackage{amssymb}
\usepackage{amsmath}
\usepackage{graphicx}
\usepackage{epstopdf}
\usepackage{amsfonts}%
\usepackage{indentfirst}
\usepackage{array}
\usepackage{color}
\usepackage{wrapfig}
\usepackage{booktabs}

\newtheorem{problem}{\textbf{Problem}}

\title[BarrierNet]{BarrierNet: A Safety-Guaranteed Layer for Neural Networks}
\usepackage{times}

 \author{%
 \Name{Wei Xiao} \Email{weixy@mit.edu}
 \AND
 \Name{Ramin Hasani} \Email{rhasani@mit.edu}
 \AND
 \Name{Xiao Li} \Email{xiaoli@mit.edu}
 \AND
 \Name{Daniela Rus}\thanks{All authors are with CSAIL MIT, Cambridge, MA, USA. }\Email{rus@mit.edu}
 }

\begin{document}

\maketitle

\begin{abstract}%
This paper introduces differentiable higher-order control barrier functions (CBF) that are end-to-end trainable together with learning systems. CBFs are usually overly conservative, while guaranteeing safety. Here, we address their conservativeness by softening their definitions using environmental dependencies without loosing safety guarantees, and embed them into differentiable quadratic programs. These novel safety layers, termed a BarrierNet, can be used in conjunction with any neural network-based controller, and can be trained by gradient descent. BarrierNet allows the safety constraints of a neural controller be adaptable to changing environments. We evaluate them on a series of control problems such as traffic merging and robot navigations in 2D and 3D space, and demonstrate their effectiveness compared to state-of-the-art approaches.


\end{abstract}

\begin{keywords}%
  Neural Network, Safety Guarantees, Control Barrier Function
\end{keywords}

\section{Introduction}
\noindent The deployment of learning systems in decision-critical applications such as autonomous ground and aerial vehicle control is strictly conditional to their safety properties. This is because one simple mistake made by a learned controller, can lead to catastrophic outcomes. Notwithstanding, ensuring the safety (e.g., providing guarantees that a self-driving car will never collide with obstacles) of modern learning-based autonomous controllers is challenging. This is because these models perform representation learning end-to-end in unstructured environments, and are expected to generalize well to unseen situations. 

In this work, we take insights from designing data-driven safety controllers \citep{ames2014control,ames2017control} to equip end-to-end learning systems with safety guarantees. 
To this end, we propose a fundamental algorithm to synthesize safe neural controllers end-to-end, by defining novel instances of control barrier functions (CBFs), that are differentiable. CBFs are popular methods for guaranteeing safety when the system dynamics are known. A large body of work studied variants of CBFs \citep{nguyen2016exponential,wang2018safe,taylor2020learning,choi2020reinforcement,taylor2020control}, and their characteristics under increasing uncertainty \citep{xu2015robustness,gurriet2018towards,taylor2020adaptive}. The common observation is that as the uncertainty of models increase CBFs make the system's behavior excessively conservative \citep{csomay2021episodic}.

In the present study, we address this over-conservativeness of CBFs by replacing the set of hard constraints in high order CBFs (HOCBFs) \citep{Xiao2021TAC2} for arbitrary-relative-degree systems, with a set of differentiable soft constraints without loss of safety guarantees. This way, we obtain a versatile safety guaranteed barrier layer, termed a BarrierNet, that can be combined with any deep learning system \citep{hochreiter1997long,he2016deep,vaswani2017attention,lechner2020learning,hasani2021closed}, and can be trained end-to-end via reverse-mode automatic differentiation \citep{rumelhart1986learning}. 

\begin{figure}[t]
	\centering
	\includegraphics[scale=1.3]{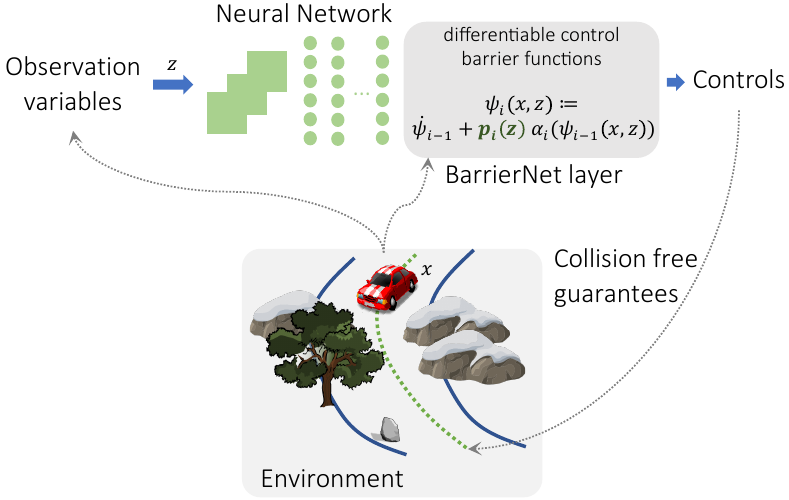}
	\caption{A safety guaranteed BarrierNet controller for autonomous driving. Collision avoidance is guaranteed. $\bm x$ is the vehicle state, and $\bm z$ is the observation variable.  $\psi_i(\bm x,\bm z), i\in\{1,\dots,m\}$ are a sequence of CBFs with their class $\mathcal{K}$ funtions $\alpha_i$ in a HOCBF with relative degree $m$. $p_i(\bm z), i\in\{1,\dots,m\}$ are trainable parameters or from the previous layer. A standard HOCBF does not have the trainable terms $p_i(\bm z)$, and thus, each $\psi_i$ is independent of the observation variable $\bm z$. }	
	\label{fig:Nctrl}
\end{figure}

BarrierNet allows the safety constraints of a neural controller to be adaptable to changing environments. A typical application of BarrierNet is autonomous driving that is shown in Fig. \ref{fig:Nctrl}. In this example, a neural network outputs acceleration and steering commands to navigate the vehicle along the center lane while avoiding obstacles. The system and environmental observations are inputs to the upstream network whose outputs serve as arguments to the BarrierNet layer. Finally, BarrierNet outputs the controls that guarantee collision avoidance. In contrast to existing work, our proposed architecture is end-to-end trainable including the BarrierNet for neural networks.

\noindent \textbf{Contributions:} $(i)$ We propose a novel trainable and interpretable layer, built by leveraging the definition of higher order control barrier functions,that provides safety guarantees for general control problems with neural network controllers. $(ii)$ We resolve the over-conservativeness of CBFs by introducing differentiable soft constraints in their definition (c.f., Fig. \ref{fig:Nctrl}). This way we can train them end-to-end with a given learning system. $(iii)$ We design BarrierNet such that the CBF parameters are adaptive to changes in environmental conditions. CBF parameters can be learned from data. We evaluate BarrierNet on a set of control problems including traffic merging and robot navigation with obstacles in 2D and 3D.

\noindent This paper is organized as follows: In Sec. \ref{sec:relate} and \ref{sec:bkg}, we provide the related work and the necessary background to construct our theory, respectively. We formulate our problem in Sec. \ref{sec:prob}  and introduce BarrierNets in Sec. \ref{sec:BN}. Sec. \ref{sec:simulation} includes our experimental evaluation and we conclude the paper in Sec. \ref{sec:conc}.



\section{Related Work} 
\label{sec:relate}

\noindent \textbf{Control Barrier Function and Learning.} A large body of work studied CBF-based safety guarantees \citep{nguyen2016exponential,wang2018safe,taylor2020learning,choi2020reinforcement,taylor2020control}, and their characteristics under increasing model uncertainty \citep{xu2015robustness,gurriet2018towards,taylor2020adaptive}. Many existing works \citep{ames2017control,nguyen2016exponential,Yang2019}
combine CBFs for systems with quadratic costs to form optimization problems. Time is discretized and an optimization problem with
constraints given by the CBFs (inequalities of the form (\ref{eqn:constraint}
)) which are solved at each time step. The inter-sampling effect is considered in \citep{Yang2019,Xiao2021CDC}. Replacing CBFs by HOCBFs allows us to handle constraints with arbitrary relative degree \citep{Xiao2021TAC2}. The common observation is that as the uncertainty of models increases, CBFs make the system's behavior excessively conservative \citep{csomay2021episodic}.

The recently proposed adaptive CBFs (AdaCBFs) \citep{Xiao2021TAC1} addressed the conservativeness of the HOCBF method by multiplying the class $\mathcal{K}$ functions of an HOCBF with some penalty functions. These penalties functions, themselves, are HOCBFs such that they are guaranteed to be non-negative. This is due to the fact that the main conservativeness of the HOCBF method comes from the class $\mathcal{K}$ functions. By multiplying (relaxing) the class $\mathcal{K}$ functions with some penalty functions, it has been shown that the satisfaction of the AdaCBF constraint is a necessary and sufficient condition for the satisfaction of the original safety constraint $b(\bm x)\geq 0$ \citep{Xiao2021TAC1}. This is conditioned on designing proper auxiliary dynamics for all the penalty functions, based on specific problems. However, how to design such auxiliary dynamics is still a remaining challenge, which we study here.

At the intersection of CBFs and learning, supervised learning techniques have been proposed to learn safe set definitions from demonstrations \citep{Robey2020}, and sensor data \citep{Srinivasan2020} which are then enforced by CBFs. \citep{taylor2020learning} used data to learn system dynamics for CBFs. In a similar setting, \citep{Lopez2020} used adaptive control approaches to estimate the unknown system parameters in CBFs.  In \citep{Yaghoubi2020}, neural network controllers are trained using CBFs in the presence of disturbances. These prior works focus on learning safe sets and dynamics, whereas we focus on the design of environment dependent and trainable CBFs.

\noindent \textbf{Optimization-based safety frameworks.} Recent advances in differentiable optimization methods show promising directions for safety guaranteed neural network controllers \citep{lechner2020gershgorin,gruenbacher2020lagrangian,grunbacher2021verification,gruenbacher2021gotube,massiani2021exploration,lechner2021adversarial}. In \citep{Amos2017}, a differentiable quadratic program (QP) layer, called OptNet, was introduced. The OptNet has been used in neural networks as a filter for safe controls \citep{pereira2020}, in which case the OptNet is not trainable, thus, could limit the system's learning performance. In contrast, we propose a trainable safety-guaranteed neural controller. In \citep{Deshmukh2019,Jin2020,Zhao2021}, safety guaranteed neural network controllers have been learned through verification-in-the-loop training. A safe neural network filter has been proposed in \citep{Ferlez2020} for a specific vehicle model using verification methods. The verification approaches can not ensure coverage of the entire state space, they are offline and are not adaptive to environment changes (such as varying size of the unsafe sets) \citep{li2021comparison}. In comparison, BarrierNet is online, easily scalable, general to control problems and is adaptive to the environment changes.

\section{Backgrounds}
\label{sec:bkg}
\noindent In this section, we briefly introduce control barrier functions (CBF) and refer interested readers to \citep{ames2017control} for detailed formulations. Intuitively, CBF is a means to translate state constraints to control constraints under affine dynamics. The controls that satisfy those constraints can be efficiently solved for by formulating a quadratic program. We start with the definition of a class $\mathcal{K}$ function.

\begin{definition}
	\label{def:classk} (\textit{Class $\mathcal{K}$ function} \citep{Khalil2002}) A
	continuous function $\alpha:[0,a)\rightarrow[0,\infty), a > 0$ is said to
	belong to class $\mathcal{K}$ if it is strictly increasing and $\alpha(0)=0$. A continuous function $\beta:\mathbb{R}\rightarrow\mathbb{R}$ is said to belong to extended class $\mathcal{K}$ if it is strictly increasing and $\beta(0)=0$.
\end{definition}

\noindent Consider an affine control system of the form
\begin{equation}
\dot{\bm{x}}=f(\bm x)+g(\bm x)\bm u \label{eqn:affine}%
\end{equation}
where $\bm x\in\mathbb{R}^{n}$, $f:\mathbb{R}^{n}\rightarrow\mathbb{R}^{n}$
and $g:\mathbb{R}^{n}\rightarrow\mathbb{R}^{n\times q}$ are {locally}
Lipschitz, and $\bm u\in U\subset\mathbb{R}^{q}$, where $U$ denotes a control constraint set.

\begin{definition}
	\label{def:forwardinv} A set $C\subset\mathbb{R}^{n}$ is forward invariant for
	system (\ref{eqn:affine}) if its solutions for some $\bm u\in U$ starting at any $\bm x(0) \in C$ satisfy $\bm x(t)\in C,$ $\forall t\geq0$.
\end{definition}

\begin{definition}
	\label{def:relative} (\textit{Relative degree}) The relative degree of a
	(sufficiently many times) differentiable function $b:\mathbb{R}^{n}%
	\rightarrow\mathbb{R}$ with respect to system (\ref{eqn:affine}) is the number
	of times it needs to be differentiated along its dynamics until the control
	$\bm u$ explicitly shows in the corresponding derivative.
\end{definition}

\noindent Since function $b$ is used to define a (safety) constraint $b(\bm
x)\geq0$, we will also refer to the relative degree of $b$ as the relative
degree of the constraint. For a constraint $b(\bm x)\geq0$ with relative
degree $m$, $b:\mathbb{R}^{n}\rightarrow\mathbb{R}$, and $\psi_{0}(\bm
x):=b(\bm x)$, we define a sequence of functions $\psi_{i}:\mathbb{R}%
^{n}\rightarrow\mathbb{R},i\in\{1,\dots,m\}$:
\begin{equation}
\begin{aligned} \psi_i(\bm x) := \dot \psi_{i-1}(\bm x) + \alpha_i(\psi_{i-1}(\bm x)),\quad i\in\{1,\dots,m\}, \end{aligned} \label{eqn:functions}%
\end{equation}
where $\alpha_{i}(\cdot),i\in\{1,\dots,m\}$ denotes a $(m-i)^{th}$ order
differentiable class $\mathcal{K}$ function.

We further define a sequence of sets $C_{i}, i\in\{1,\dots,m\}$ associated
with (\ref{eqn:functions}) in the form:
\begin{equation}
\label{eqn:sets}\begin{aligned} C_i := \{\bm x \in \mathbb{R}^n: \psi_{i-1}(\bm x) \geq 0\}, \quad i\in\{1,\dots,m\}. \end{aligned}
\end{equation}

\begin{definition}
	\label{def:hocbf} (\textit{High Order Control Barrier Function (HOCBF)}
	\citep{Xiao2021TAC2}) Let $C_{1}, \dots, C_{m}$ be defined by (\ref{eqn:sets}%
	) and $\psi_{1}(\bm x), \dots, \psi_{m}(\bm x)$ be defined by
	(\ref{eqn:functions}). A function $b: \mathbb{R}^{n}\rightarrow\mathbb{R}$ is
	a High Order Control Barrier Function (HOCBF) of relative degree $m$ for
	system (\ref{eqn:affine}) if there exist $(m-i)^{th}$ order differentiable
	class $\mathcal{K}$ functions $\alpha_{i},i\in\{1,\dots,m-1\}$ and a class
	$\mathcal{K}$ function $\alpha_{m}$ such that
	\begin{equation}
	\label{eqn:constraint}\begin{aligned} 
	\sup_{\bm u\in U}[L_f^{m}b(\bm x) + [L_gL_f^{m-1}b(\bm x)]\bm u \!+\! O(b(\bm x)) + \alpha_m(\psi_{m-1}(\bm x))] \geq 0, \end{aligned}
	\end{equation}
	for all $\bm x\in C_{1} \cap,\dots, \cap C_{m}$. In
	(\ref{eqn:constraint}), $L_{f}^{m}$ ($L_{g}$) denotes Lie derivatives along
	$f$ ($g$) $m$ (one) times, and $O(b(\bm x)) = \sum_{i = 1}^{m-1}L_f^i(\alpha_{m-i}\circ\psi_{m-i-1})(\bm x).$ {Further, $b(\bm x)$ is such that $L_gL_f^{m-1}b(\bm x)\ne 0$ on the boundary of the set $C_{1} \cap,\dots, \cap C_{m}$.}
\end{definition}

The HOCBF is a general form of the relative degree one CBF
\citep{ames2017control} (setting $m=1$ reduces the HOCBF to
the common CBF form). {Note that we can define $\alpha_i(\cdot), i\in\{1,\dots, m\}$ in Def. \ref{def:hocbf} to be extended class $\mathcal{K}$ functions to ensure robustness of a HOCBF to perturbations \citep{ames2017control}. However, the use of extended class $\mathcal{K}$ functions cannot ensure a constraint is eventually satisfied if it is initially violated.}

\begin{theorem}
	\label{thm:hocbf} (\citep{Xiao2021TAC2}) Given a HOCBF $b(\bm x)$ from Def.
	\ref{def:hocbf} with the associated sets $C_{1}, \dots, C_{m}$ defined
	by (\ref{eqn:sets}), if $\bm x(0) \in C_{1} \cap,\dots,\cap C_{m}$,
	then any Lipschitz continuous controller $\bm u(t)$ that satisfies
	the constraint in (\ref{eqn:constraint}), $\forall t\geq0$ renders $C_{1}\cap,\dots,
	\cap C_{m}$ forward invariant for system (\ref{eqn:affine}).
\end{theorem}

\noindent We provide a summary of notations in Table \ref{tab:1}.

\begin{table}[t]
    \centering
    \caption{Notation}\label{tab:1}
    \begin{tabular}{cc}
    \toprule
       \textbf{Symbol}  & \textbf{Definition} \\
         \midrule 
         $t$ & time \\
        $\theta$ & BarrierNet parameters \\
         $\bm x \in \mathbb{R}^n$ &  vehicle state\\
         $\bm z \in \mathbb{R}^d$ &  observation variable\\
         $\bm u \in \mathbb{R}^q$ &  control \\
         $\alpha: [0,a)\rightarrow [0, \infty), a>0$ & class $\mathcal{K}$ function \\
         $\beta: \mathbb{R} \rightarrow \mathbb{R}$ & extended class $\mathcal{K}$ function \\
         $b: \mathbb{R}^n \rightarrow \mathbb{R}$ & safety constraint \\
         $\psi: \mathbb{R}^n \rightarrow \mathbb{R}$ &  CBF \\
         $p: \mathbb{R}^d \rightarrow \mathbb{R}^{>0}$ &  penalty function\\
         $C \subset \mathbb{R}^n$ &  safe set\\
         $l: \mathbb{R}^q \times \mathbb{R}^q \rightarrow \mathbb{R}$ &  similarity measure\\
         $f: \mathbb{R}^n \rightarrow \mathbb{R}^n$ & affine dynamics drift term \\
         $g: \mathbb{R}^n \rightarrow \mathbb{R}^{n \times q}$ & affine dynamics control term \\
    \bottomrule
    \end{tabular}
    \label{tab:notation}
\end{table}

\section{Problem Formulation}
\label{sec:prob}
\noindent Here, we formally define the learning problem for safety-critical control. 

\begin{problem}\label{prob:1}
Given (i) a system with known affine dynamics in the form of \ref{eqn:affine}, (ii) a nominal controller $f^\star(\bm x)=\bm u^\star$, (iii) a set of safety constraints $b_j(\bm x) \geq 0, j\in S$ (where $b_j$ is continuously differentiable), (iv) control bounds $\bm u_{min}\leq\bm u\leq\bm u_{max}$ and (v) a neural network controller $f(\bm x | \theta)=\bm u$ parameterized by $\theta$, our goal is to find the optimal parameters 

\begin{equation}
    \theta^\star = \underset{\theta}{\arg\min}\mathbb{E}_{\bm x}[l(f^\star(\bm x), f(\bm x|\theta))]
\end{equation}

\noindent while guaranteeing the satisfaction of the safety constraints in (iii) and control bounds in (iv). $\mathbb{E}(\cdot)$ is the expectation and $l(\cdot, \cdot)$ denotes a similarity measure.

\end{problem}

Problem \ref{prob:1} defines a policy distillation problem with safety guarantees. The safety constraints can be pre-defined by users or can be learned from data \citep{Robey2020}, \citep{Srinivasan2020}.






\section{BarrierNet}
\label{sec:BN}

\noindent In this section, we propose BarrierNet - an CBF-based neural network controller with parameters that are trainable via backpropagation.  We define the safety guarantees of a neural network controller as follows:
\begin{definition}
	(Safety guarantees) A neural network controller has safety guarantees for system (\ref{eqn:affine}) if its outputs (controls) drive system (\ref{eqn:affine}) such that $b_j(\bm x(t)) \geq 0, \forall t\geq 0, \forall j\in S$, for continuously differentiable $b_j:\mathbb{R}^n\rightarrow \mathbb{R}$.
\end{definition}

We start with the motivations of a BarrierNet. The HOCBF method provides safety guarantees for control systems with arbitrary relative degrees, but in a conservative way. In other words, the satisfaction of the HOCBF constraint (\ref{eqn:constraint}) is only a sufficient condition of the satisfaction of the original safety constraint $b(\bm x)\geq 0$. This conservativeness of the HOCBF method will significantly limit the system performance. For example, such conservativeness may drive the system much further away from obstacles than necessary. Our first motivation is to address this conservativeness.

More specifically, a HOCBF constraint is always a hard constraint in order to guarantee safety. This may adversely affect the performance of the system. In order to address this, we soften a HOCBF in a way without loosing the safety guarantees. In addition, we also incorporate HOCBF in a differentiable optimization layer to allow tuning of its parameters from data. 

Given a safety requirement $b(\bm x)\geq 0$ with relative degree $m$ for system (\ref{eqn:affine}), we change the sequence of CBFs in (\ref{eqn:functions}) by:
\begin{equation}
\begin{aligned} \psi_i(\bm x, \bm z) := \dot \psi_{i-1}(\bm x, \bm z) + p_i(\bm z)\alpha_i(\psi_{i-1}(\bm x, \bm z)),\quad i\in\{1,\dots,m\}, \end{aligned} \label{eqn:layers}%
\end{equation}
where $\psi_0(\bm x, \bm z) = b(\bm x)$ and $\bm z \in \mathbb{R}^d$ are the input of the neural network ($d\in\mathbb{N}$ is the dimension of the features), $p_i:\mathbb{R}^d\rightarrow \mathbb{R}^{>0}, i\in \{1,\dots,m\}$ are the outputs of the previous layer, where $\mathbb{R}^{>0}$ denotes the set of positive scalars. The above formulation is similar to the Adaptive CBF (AdaCBF) \citep{Xiao2021TAC1}, but is trainable, and does not require us to design auxiliary dynamics for $p_i$ (an AdaCBF does require), a non-trivial process of the existing AdaCBF method. In order to make sure that the HOCBF method can be solved in a time-discretization way \citep{ames2017control}, we require that each $p_i$ is Lipschitz continuous. Then, we have a similar HOCBF constraint as in Def. \ref{def:hocbf} in the form:
\begin{equation}
	\label{eqn:NCBF}\begin{aligned} 
	L_f^{m}b(\bm x) + [L_gL_f^{m-1}b(\bm x)]\bm u \!+\! O(b(\bm x), \bm z) + p_m(\bm z)\alpha_m(\psi_{m-1}(\bm x, \bm z)) \geq 0, \end{aligned}
\end{equation}

Following the discretization solving method introduced in \citep{ames2017control}, if we wish to minimize the control input effort (in the case where the reference control is 0), we can reformulate our problem as the following sequence of QPs:
\begin{equation} \label{eqn:obj}
\bm u^*(t) = \arg\min_{\bm u(t)} \frac{1}{2}\bm u(t)^TH\bm u(t)
\end{equation}
s.t.
$$
\begin{aligned} 
	L_f^{m}b_j(\bm x) + [L_gL_f^{m-1}b_j(\bm x)]\bm u \!+\! O(b_j(\bm x), \bm z) + p_m(\bm z)\alpha_m(\psi_{m-1}(\bm x, \bm z)) \geq 0, j\in S \end{aligned}
$$
$$
\bm u_{min}\leq\bm u\leq\bm u_{max},
$$
$$
t = k\Delta t + t_0,
$$


\noindent where $\Delta t > 0$ is the discretization time. Since the QPs in (\ref{eqn:obj}) is solved pointwise, the resulting solutions would be sub-optimal.  In order to address this problem and be able to track any nominal controllers, we introduce BarrierNet.

\begin{definition}\label{def:neuron}
    (BarrierNet) A BarrierNet is composed by neurons in the form:
\begin{equation} \label{eqn:neuron}
\bm u^*(t) = \arg\min_{\bm u(t)} \frac{1}{2}\bm u(t)^TH(\bm z | \theta_h)\bm u(t) + F^T(\bm z | \theta_f)\bm u(t)
\end{equation}
s.t.
\begin{equation}\label{eqn:10}
\begin{aligned} 
	L_f^{m}b_j(\bm x) + [L_gL_f^{m-1}b_j(\bm x)]\bm u \!+\! O(b_j(\bm x), \bm z| \theta_p) + p_m(\bm z | \theta^m_p)\alpha_m(\psi_{m-1}(\bm x, \bm z|\theta_p)) \geq 0, j\in S \end{aligned}
\end{equation}
$$
\bm u_{min}\leq\bm u\leq\bm u_{max},
$$
$$
t = k\Delta t + t_0,
$$
where $F(\bm z | \theta_f)\in\mathbb{R}^q$ could be interpreted as a reference control (can be the output of previous network layers) and $\theta_h, \theta_f, \theta_p = (\theta_p^1, \dots, \theta_p^m)$ are trainable parameters.
\end{definition} 

\noindent The inequality (\ref{eqn:10}) in Definition \ref{def:neuron} guarantees each safety constraint $b_j(\bm x)\geq 0, \forall j\in S$ through the parameterized function $p_i, i\in\{1,\dots, m\}$. Based on Def. \ref{def:neuron}, for instance, if we have 10 control agents, we need 10 BarrierNet neurons presented by (\ref{eqn:neuron}) to ensure the safety of each agent. This implies that BarrierNets can be extended to multi-agent settings.



In Def. \ref{def:neuron}, we make both $H(\bm z | \theta_h)$ and $F(\bm z | \theta_f)$ parameterized and dependent on the network input $\bm z$, but $H$ and $F$ can also be directly trainable parameters that do not depend on the previous layer (i.e., we have $H$ and $F$). The same applies to $p_i, i\in\{1,\dots, m\}$. The trainable parameters are $\theta = \{\theta_h, \theta_f, \theta_p\}$ (or $\theta = \{H, F, p_i, \forall i\in\{1,\dots,m\}\}$ if $H, F$ and $p_i$ do not depend on the previous layer). The solution $\bm u^*$ is the output of the neuron. The BarrierNet is differentiable with respect to its parameters \citep{Amos2017}. We describe the forward and backward passes as follows.



\textbf{Forward pass:} The forward step of a BarrierNet is to solve the QP in Definition \ref{def:neuron}. The inputs of a BarrierNet include environmental features $\bm z$ (such as the location and speed of an obstacle) that can be provided directly or from a tracking network if raw sensory inputs are used. BarrierNet also takes as inputs the system states $\bm x$ as a feedback, as shown in Fig. \ref{fig:Nctrl}. The outputs are the solutions of the QP (the resultant controls). 

\textbf{Backward pass:} The main task of BarrierNet is to provide controls while always ensuring safety. Suppose $\ell$ denotes some loss function (a similarity measure). Using the techniques introduced in \citep{Amos2017}, the relevant gradient with respect to all the BarrierNet parameters can be given by  (the gradient with respect to the parameters $\theta_h, \theta_f, \theta_p$ can be obtained using the chain rule):
\begin{equation} \label{eqn:loss}
\begin{aligned}
    &\bigtriangledown_{H} \ell = \frac{1}{2}(d_{\bm u}\bm u^T + \bm ud^T_{\bm u}), \qquad \bigtriangledown_{F} \ell = d_{\bm u},\\
    &\bigtriangledown_{G} \ell = D(\lambda^*)(d_{\lambda}\bm u^T + \lambda d_{\bm u}^T), \; \bigtriangledown_{h} \ell = -D(\lambda^*)d_{\lambda},
\end{aligned}
\end{equation}
where $\lambda$ are the dual variables on the HOCBF constraints and $D(\cdot)$ creates diagonal matrix from a vector. $G, h$ are concatenated by $G_j, h_j, j\in S$, where 
\begin{equation} \label{eqn:concat}
\begin{aligned}
    G_j &= -L_gL_f^{m-1}b_j(\bm x),\\
    h_j &= 	L_f^{m}b_j(\bm x)  \!+\! O(b_j(\bm x), \bm z) + p_m(\bm z)\alpha_m(\psi_{m-1}(\bm x, \bm z)).
\end{aligned}    
\end{equation}
Since the control bounds in (\ref{eqn:neuron}) are not trainable, they are not included in $G, h$.

In (\ref{eqn:loss}), $d_{\bm u}$ and $d_{\lambda}$ are given by solving:
\begin{equation}
\left[
\begin{array}
[c]{c}%
d_{\bm u}\\
d_{\lambda}
\end{array}
\right]  = \left[
\begin{array}
[c]{cc}%
H & G^TD(\lambda^*)\\
G & D(G\bm u^* - h)
\end{array}
\right]^{-1} \left[\begin{array}{c}  
	(\frac{\partial\ell}{\partial \bm u^*})^T\\
	0
	\end{array} \right],  %
\end{equation}

The above equation is obtained by taking the Lagrangian of QP followed by applying the Karush–Kuhn–Tucker conditions. $\bigtriangledown_{G} \ell$ is not applicable in a BarrierNet as it is determined by the corresponding HOCBF. $\bigtriangledown_{h} \ell$ is also not directly related to the input of a BarrierNet. Nevertheless, we have $\bigtriangledown_{p_i} \ell = \bigtriangledown_{h_j} \ell \bigtriangledown_{p_i}h_j, i\in\{1,\dots,m\}, j\in S$ ,where $\bigtriangledown_{h_j} \ell$ is given by $\bigtriangledown_{h} \ell$ in (\ref{eqn:loss}) and $\bigtriangledown_{p_i} h_j$ is given by taking the partial derivative of $h_j$ in (\ref{eqn:concat}).

The Following theorem characterizes the safety guarantees of a BarrierNet:

\begin{theorem}
    If $p_i(\bm z), i\in\{1,\dots, m\}$ are Lipschitz continuous, then a BarrerNet composed by neurons as in Def. \ref{def:neuron} guarantees the safety of system (\ref{eqn:affine}).
\end{theorem}
\textbf{proof:} Since $p_i(\bm z), i\in\{1,\dots, m\}$ are Lipschitz continuous, it follows from Thm. \ref{thm:hocbf} that each $\psi_i(\bm x, \bm z)$ in (\ref{eqn:layers}) is a valid CBF. Starting from $\psi_m(\bm x,\bm z)\geq 0$ (the non-Lie-derivative form of each HOCBF constraint  (\ref{eqn:NCBF})), we can show that $\psi_{m-1}(\bm x,\bm z)\geq 0$ is guaranteed to be satisfied following Thm. \ref{thm:hocbf}. Recursively, we can show that $\psi_0(\bm x, \bm z)\geq 0$ is guaranteed to be satisfied. As $b(\bm x) = \psi_0(\bm x, \bm z)$ following (\ref{eqn:layers}), we have that system (\ref{eqn:affine}) is safety guaranteed in a BarrierNet. $\blacksquare$

\begin{remark}
(Adaptivity of the BarrierNet) The HOCBF constraints in a BarrierNet are softened by the trainable penalty functions without loosing safety guarantees. The penalty functions are environment dependent which features can be calculated from upstream networks. The adaptive property of the HOCBFs provides the adaptivity of the BarrierNet. As a result, BarrierNet is able to generate safe controls while avoid being overly conservative.
\end{remark}

In order to guarantee that $p_i(\bm z), i\in\{1,\dots, m\}$ are Lipschitz continuous, we can choose activation functions of the previous layer as some continuously differentiable functions, such as sigmoid functions. The process of constructing and training a BarrierNet includes: $(a)$ construct a softened HOCBF by (\ref{eqn:layers}) that enforces each of the safety requirement, $(b)$ construct the parameterized BarrierNet by (\ref{eqn:neuron}), (c) get the training data set using the nominal controller, and $(d)$ train the BarrierNet using error backpropogation. We summarize the algorithm for the BarrierNet in Alg. \ref{alg:et}.


\begin{algorithm}
	\caption{Construction and training of the BarrierNet} \label{alg:et}
	\KwIn{Dynamics (\ref{eqn:affine}), Safety requirements in Problem \ref{prob:1}, a nominal controller}
	\KwOut{A safety-guaranteed BarrierNet controller.}
	$\;\;$(a) Construct softened HOCBFs by (\ref{eqn:layers})\\
	(b) Construct the BarrierNet by (\ref{eqn:neuron})\\
	(c) Get the training data set using the nominal controller\\
	(d) Initialize the BarrierNet parameter $\theta$, the Epochs, and the learning rate $\gamma$\\
	\While{$e$ in Epochs}{
	$\;\;$Forward: Solve (\ref{eqn:neuron}) and get the loss $\ell$\\
	$\;\;$Backward: Get the loss gradient $\bigtriangledown_{\theta} \ell$ using (\ref{eqn:loss})\\
	$\theta \leftarrow \theta - \gamma\bigtriangledown_{\theta} \ell$
	}
	\Return $\theta$ (optimal parameters in the BarrierNet)
\end{algorithm}


\textbf{Limitations:} The proposed BarrierNet can theoretically guarantee system safety. However, there are also some limitations:

$(i)$ The number of safety constraints in a BarrierNet should be defined in training. In some scenarios, the number of safety constraints may be time-varying. Therefore, how to match multiple safety constraints with the definition of a BarrierNet remains a challenge. It is possible that we may define more than necessary constraints in a BarrierNet, and only enable those when required. 

$(ii)$ The BarrierNet is solved in discrete time, as we can only feed discrete data into the neural network. The inter-sampling effect (the system's trajectory between time intervals) should also be considered in order to achieve safety guarantees. The inter-sampling effect is sensitive at the boundary of the safety set. {Therefore, we need to avoid sampling training data at the safety set boundary. However, due to the Lyapunov property of HOCBFs, the system will always stay close to the safety set boundary if the safety constraint is violated due to the inter-sampling effect, or some perturbations. A possible approach to address the inter-sampling effect is the data-driven event-triggered framework \citep{Xiao2021CDC}.}

\section{Numerical Evaluations}
\label{sec:simulation}
\noindent In this section, we present three case studies (a traffic merging control problem and robot navigation problems in 2D and 3D) to verify the effectiveness of our proposed BarrierNet. 
\subsection{Traffic Merging Control}


\noindent\textbf{Experiment setup.} The traffic merging problem arises when traffic must be joined from two different roads, usually associated with a main lane and a merging lane as shown in Fig.1. We consider the case where all traffic consisting of controlled autonomous vehicles (CAVs) arrive randomly at the origin ($O$ and $O^\prime$) and join at the Merging Point (MP) $M$ where a lateral
collision may occur. The segment from the origin to the merging point $M$ has length $L$ for both lanes, and is called the
Control Zone (CZ). All CAVs do not overtake each other in the CZ as each road consists of a single lane. A coordinator is
associated with the MP whose function is to maintain a First-In-First-Out
(FIFO) queue of CAVs based on their arrival time at the CZ. The coordinator also enables real-time communication among the CAVs that are in the CZ including the last one leaving the CZ. The FIFO assumption, imposed so that CAVs cross the MP in
their order of arrival, is made for simplicity and often to ensure fairness.

\begin{figure}[ptbh]
	\vspace{-4mm}
\centering
\hspace*{-2mm} \includegraphics[scale=0.21]{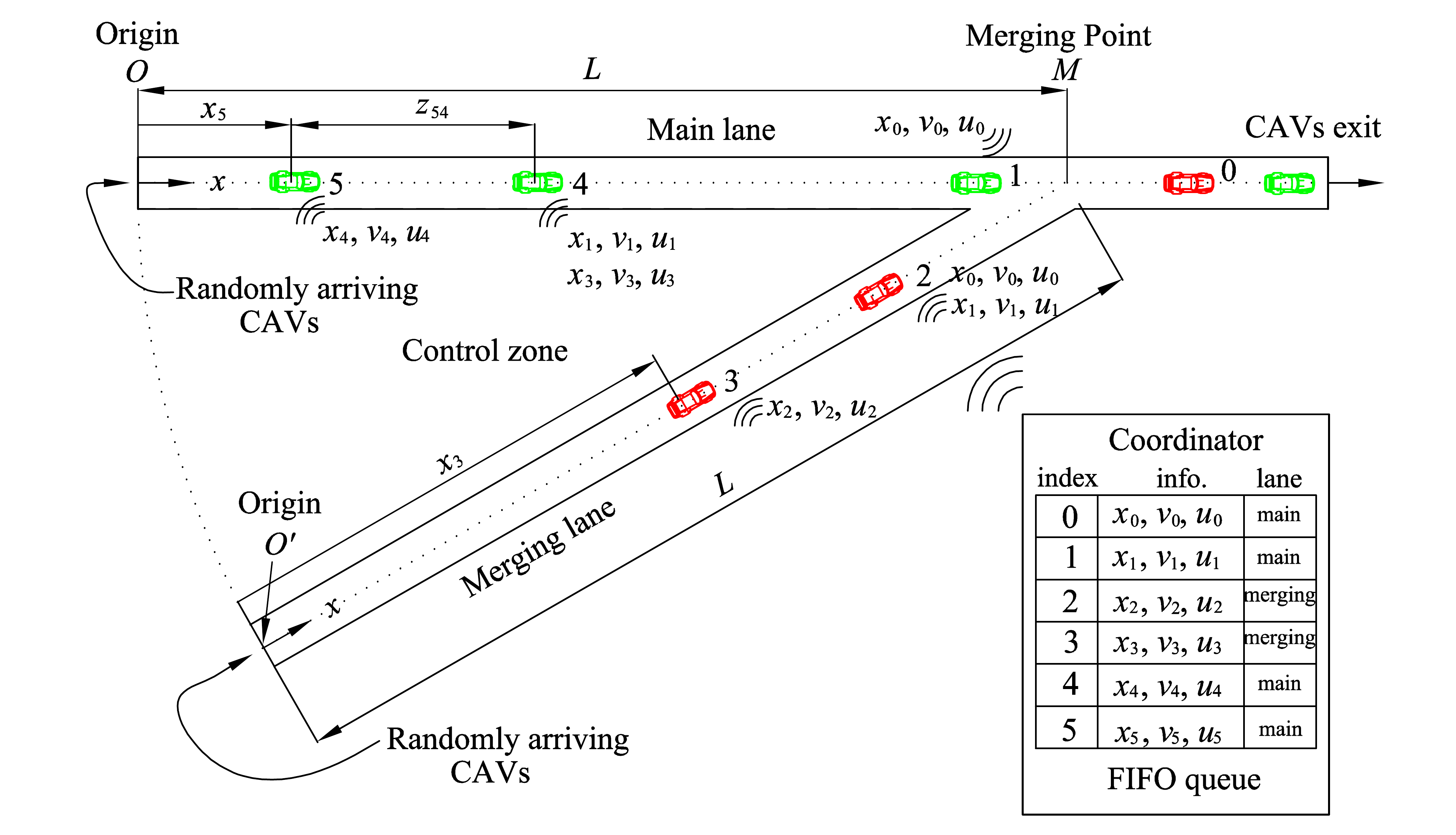} 
	\vspace{-0mm}
\caption{A traffic merging problem. A collision may happen at the merging point as well as everywhere within the control zone.}%
\label{fig:model}%
\end{figure}

\noindent\textbf{Notation.} $x_k, v_k, u_k$ denote the along-lane position, speed and acceleration (control) of CAV $k$, respectively. $t_k^0, t_k^m$ denote the arrival time of CAV $k$ at the origin and the merging point, respectively. $z_{k, k_p}$ denotes the along lane distance between CAV $k$ and its preceding CAV $k_p$, as shown in Fig. \ref{fig:model}. 

Our goal is to jointly minimize all vehicles' travel time and energy consumption in the control zone. Written as an objective function, we have
\begin{equation}
\min_{u_{k}(t)}\beta(t_{k}^{m}-t_{k}^{0})+\int_{t_{k}^{0}%
}^{t_{k}^{m}}\frac{1}{2}u_{k}^{2}(t)dt,\label{eqn:energyobj}%
\end{equation}
where $u_k$ is the vehicle's control (acceleration), and $\beta > 0$ is a weight controlling the relative magnitude of travel time and energy consumption. We assume double integrator dynamics for all vehicles.

Each vehicle $k$ should satisfy the following rear-end safety constraint if its preceding vehicle $k_p$ is in the same lane:
\begin{equation}
z_{k,k_{p}}(t)\geq\phi v_{k}(t)+\delta,\text{ \ }\forall t\in\lbrack
t_{k}^{0},t_{k}^{m}], \label{Safety}%
\end{equation}
where $z_{k,k_{p}} = x_{k_p} - x_k$ denotes the along-lane distance between $k$ and $k_p$, $\phi$ is the reaction time (usually takes $1.8s$) and $\delta \geq 0$.

The traffic merging problem is to find an optimal control that minimizes (\ref{eqn:energyobj}), subject to 
(\ref{Safety}). We assume vehicle $k$ has access to only the information of its immediate neighbors from the coordinator (shown in Fig. \ref{fig:model}), such as 
the preceding vehicle $k_p$.  This merging problem can be solved analytically by optimal control methods \citep{Xiao2021AutoA}, but at the cost of extensive computation, and the solution becomes 
complicated when one or more constraints become active in an optimal trajectory, hence possibly prohibitive 
for real-time implementation. 

\textbf{BarrierNet design.} We enforce the safety constraint (\ref{Safety}) by a CBF $b(z_{k,k_p}, v_k) = z_{k,k_{p}}(t) -\phi v_{k}(t)-\delta$, and any control input $u_k$ should satisfy the  CBF constraint (\ref{eqn:constraint}) which in this case (choose $\alpha_1$ as a linear function in Def. \ref{def:hocbf}) is :
\begin{equation}
\varphi u_k(t) \leq v_k(t) - v_{k_p}(t) + p_1(\bm z) (z_{k,k_{p}}(t) -\phi v_{k}(t)-\delta)   
\end{equation}
where $v_k$ is the speed of vehicle $k$ and $\bm z = (x_{k_p}, v_{k_p}, x_k, v_k)$ is the input of the neural network model (to be designed later). $p_1(\bm z)$ is called a penalty in the CBF that addresses the conservativeness of the CBF method. The cost in the neuron of the BarrierNet is given by:
\begin{equation}
    \min_{u_k} (u_k - f_1(\bm z))^2
\end{equation}
where $f_1(\bm z)$ is a reference to be trained (the output of the FC network). Then,
we create a neural network model whose structure is composed by a fully connected (FC) network (an input layer and two hidden layers) followed by a BarrierNet. The input of the FC network is $\bm z$, and its output is the penalty $p_1(\bm z)$ and the reference $f_1(\bm z)$. While the input of the BarrierNet is the penalty $p_1(\bm z)$ and the reference $f_1(\bm z)$, and its output is applied to control a vehicle $k$ in the control zone. 

\begin{figure}[thpb]
	\centering
	\includegraphics[scale=0.5]{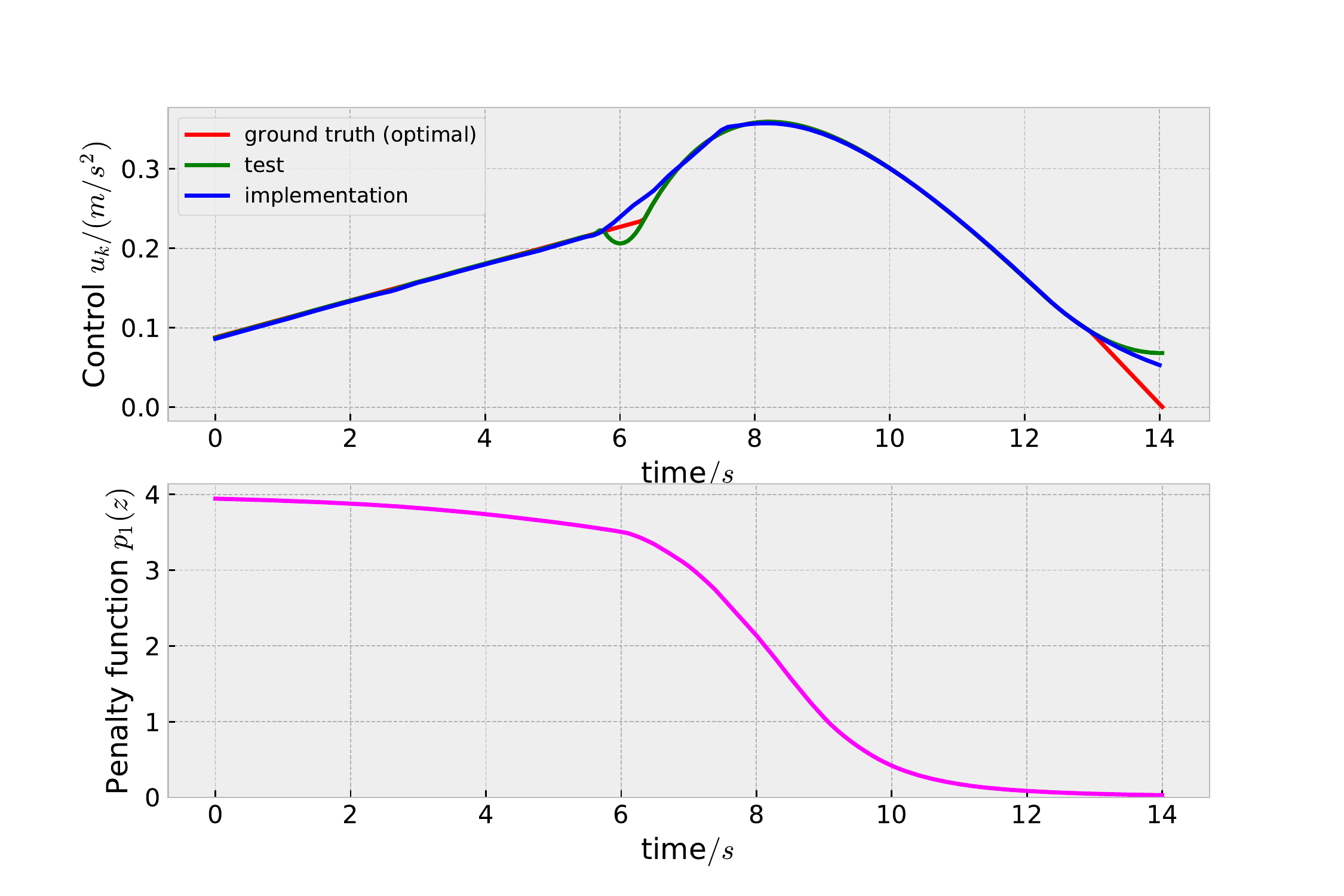}
	\caption{The control and penalty function $p_1(\bm z)$ from the BarrierNet when training with the optimal controller. The blues curves (labeled as implementation) are the vehicle control when we apply the BarrierNet to drive the vehicle dynamics to pass through the control zone. }	
	\label{fig:control_oc}
\end{figure}

\begin{figure}[htbp] 
	\centering
	
	\subfigure[Training using the optimal controller.]{
		\begin{minipage}[t]{0.45\linewidth}
			\centering
			\includegraphics[scale=0.45]{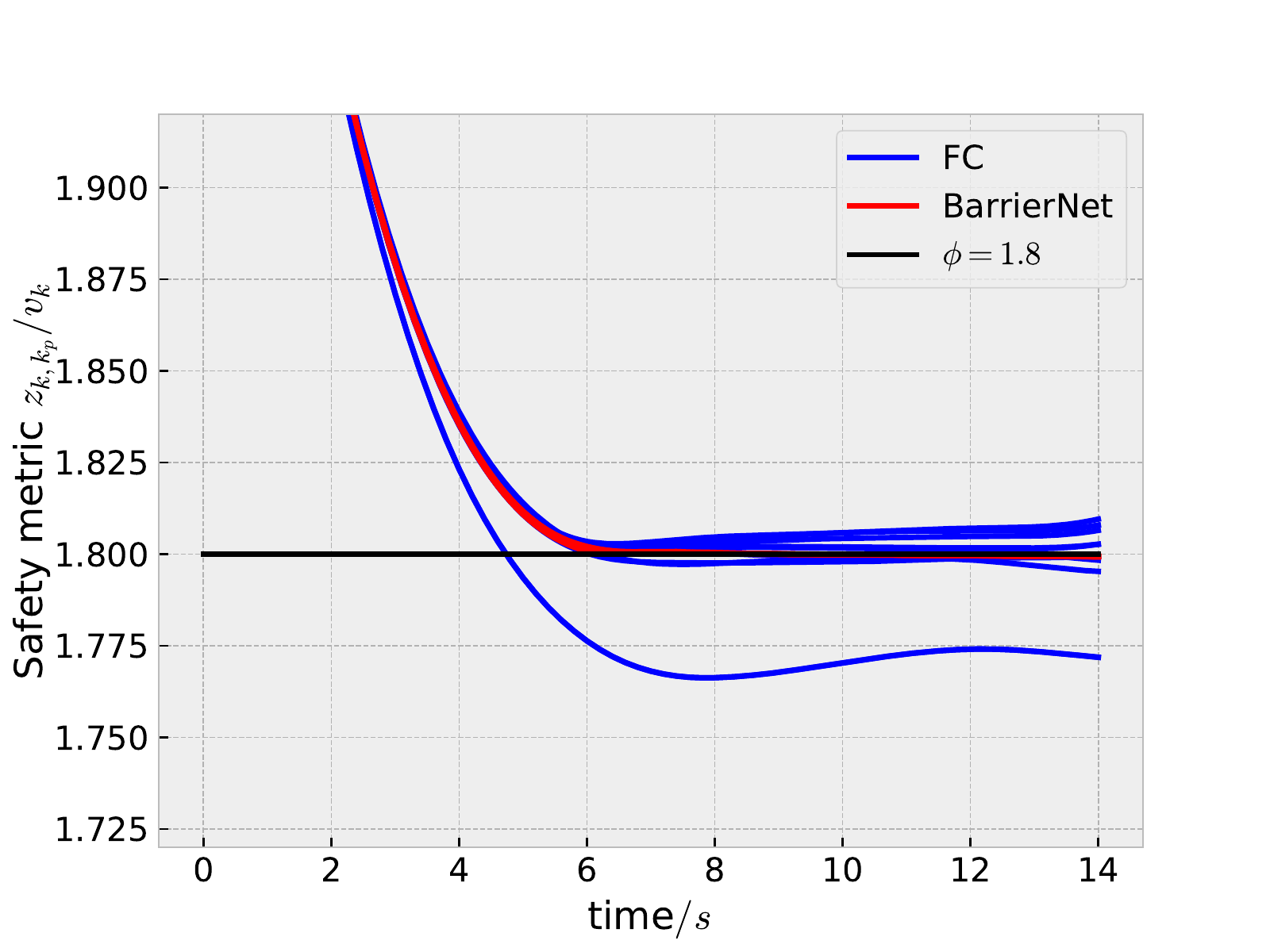} 
			\label{fig:safe}%
		\end{minipage}%
	}
	\subfigure[Training using the OCBF controller]{
		\begin{minipage}[t]{0.45\linewidth}
			\centering
			\includegraphics[scale=0.45]{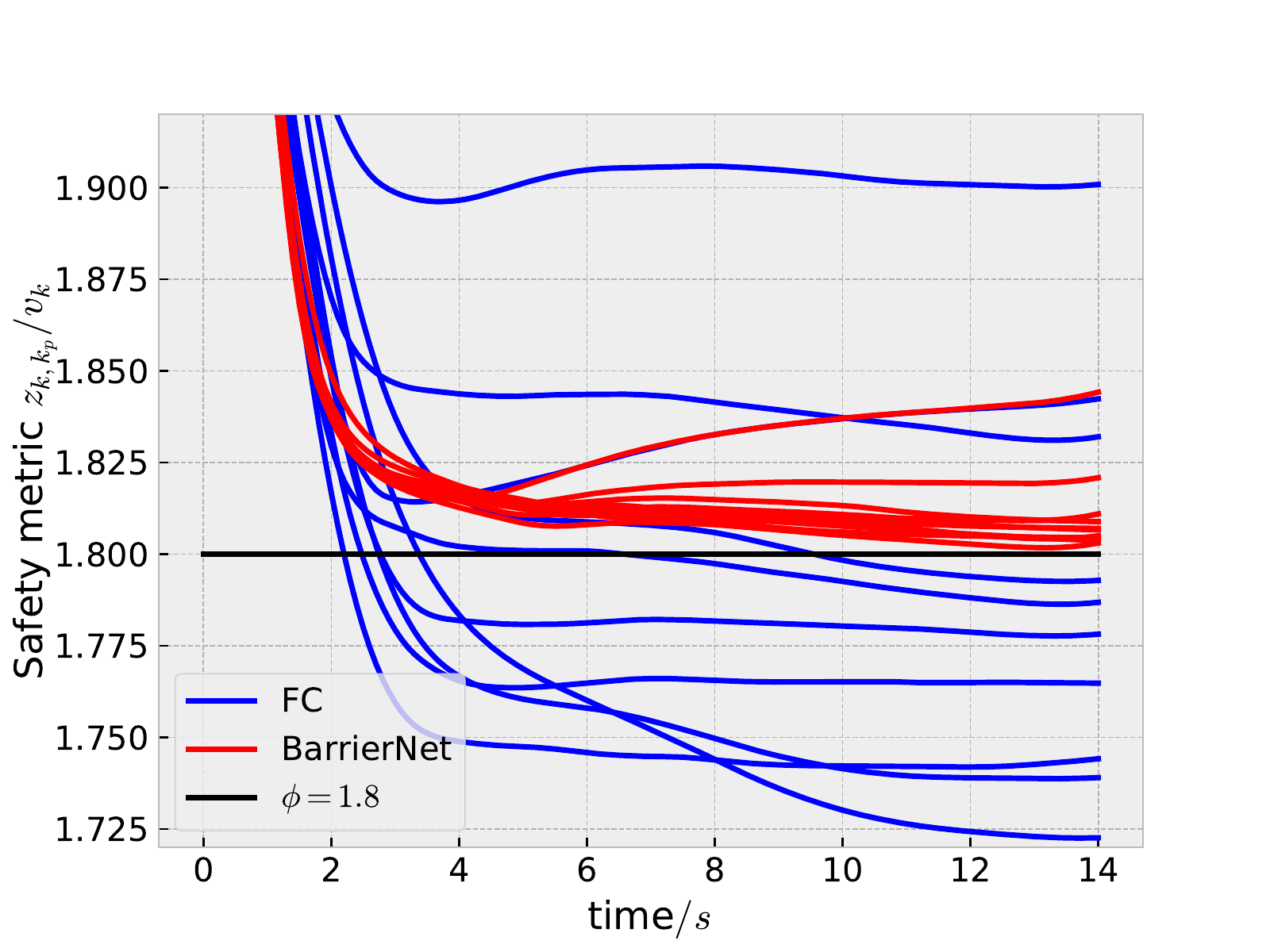} 
			\label{fig:safe_ocbf}%
		\end{minipage}%
	}
	
	\centering
	\caption{The safety comparison (under 10 trained models) between the BarrierNet and a FC network when training using the optimal/OCBF controller ($\delta = 0$). If $z_{k,k_p}/v_k$ is above the line $\phi = 1.8$, then safety is guaranteed. We observe that only neural network agents equipped with BarrierNet satisfy this condition.}
	\label{fig:safeties}
\end{figure}

\begin{figure}[thpb]
	\centering
	\includegraphics[scale=0.5]{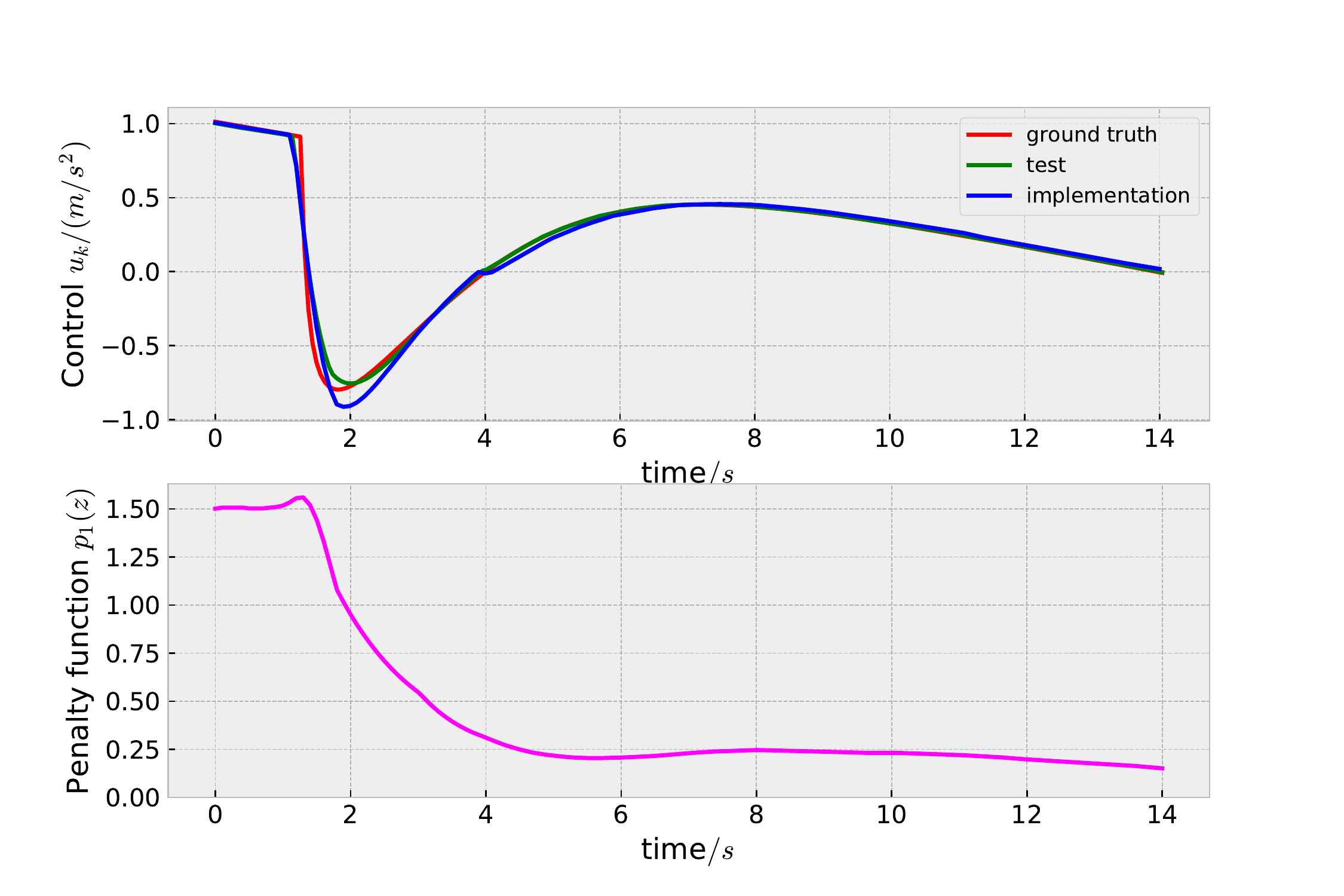}
	\caption{The control and penalty function $p_1(\bm z)$ from the BarrierNet when training with the OCBF controller. The blues curves (labeled as implementation) are the vehicle control when we apply the BarrierNet to drive the vehicle dynamics to pass through the control zone. }	
	\label{fig:control_ocbf}
\end{figure}

\textbf{Results and discussion.} To get the training data set, we solve an optimal or joint optimal control and barrier function (OCBF) \citep{Xiao2021AutoB} controller offline. The solutions of an optimal or OCBF controller are taken as labels. The training results with the optimal controller and the OCBF controller are shown in Figs. \ref{fig:control_oc} - \ref{fig:control_ocbf}.

In an optimal controller, the original safety constraint is active after around $6s$, as shown in Fig. \ref{fig:control_oc}. Therefore, the sampling trajectory is on the safety boundary, and inter-sampling effect becomes important in this case. Since we do not consider the inter-sampling effect in this paper, the safety metric of the BarrierNet might go below the lower bound $\phi =1.8$, as the red curves shown in Fig. \ref{fig:safeties}a. However, due to the Lyapunov property of the CBF, the safety metric will always stay close to the lower bound $\phi =1.8$. The solutions for 10 trained models are also consistent. In a FC network, the safety metrics vary under different trained models, and the safety constraint might be violated, as the blue curves shown in Fig. \ref{fig:safeties}a.

In an OCBF controller, the original safety constraint is not active, and thus, the inter-sampling effect is not sensitive. As shown in Figs. \ref{fig:safeties}b, safety is always guaranteed in a BarrierNet under 10 trained models. While in a FC network, the safety constraint may be violated as there are no guarantees.  

We present the penalty functions when training with the optimal controller and the OCBF controller in Figs. \ref{fig:control_oc} and \ref{fig:control_ocbf}, respectively.
The penalty function $p_1(\bm z)$ decreases when the CBF constraint becomes active. This shows the adaptivity of the BarrierNet. This behavior is similar to the AdaCBF, but in the BarrierNet, we do not need to design auxiliary dynamics for the penalty functions. Therefore, the BarrierNet is simpler than the AdaCBF. Finally, we present a comprehensive comparison between the BarrierNet, the FC network, the optimal controller and the OCBF controller in Table \ref{tab:comp}.

\begin{table}
	\caption{ Comparisons between the BarrierNet, the FC network, the optimal controller and the OCBF controller}
	\label{tab:comp}
	\centering
	\begin{tabular}{p{2.0cm}<{\centering}p{2.4cm}<{\centering}p{2.1cm}<{\centering}p{2.4cm}<{\centering}p{2cm}<{\centering}}\toprule
		item    & R.T. compute time      & safety guarantee     & Optimality  & Adaptive \\\midrule
		BarrierNet & $<0.01s$    &Yes    & close-optimal     &Yes   \\
		FC     & $<0.01s$  & No     &  close-optimal  &No     \\
		Optimal  & $30s$  & Yes &  optimal & Yes  \\
		OCBF  & $<0.01s$  & Yes &  sub-optimal & Yes  \\\bottomrule
	\end{tabular}
\end{table}

\subsection{2D Robot Navigation}


\noindent\textbf{Experiment setup.} We consider a robot navigation problem with obstacle avoidance. In this case, we consider nonlinear dynamics with two control inputs and nonlinear safety constraints. The robot navigates according to the following unicycle model for a wheeled mobile robot:
\begin{equation}\label{eqn:robot}
\left[\begin{array}{c} 
\dot x\\
\dot y\\
\dot \theta\\
\dot v
\end{array} \right]=
\left[\begin{array}{c}  
v\cos(\theta)\\
v\sin(\theta)\\
0 \\
0
\end{array} \right]  + 
\left[\begin{array}{cc}  
0 & 0\\
0 & 0\\
1 & 0\\
0 & 1
\end{array} \right]\left[\begin{array}{c}  
u_{1}\\
u_{2}
\end{array} \right]
\end{equation}
where $\bm x := (x,y,\theta,v), \bm u = (u_1,u_2)$, $x, y$ denote the robot's 2D coordinates, $\theta$ denotes the heading angle of the robot, $v$ denotes the linear speed, and $u_1, u_2$ denote the two control inputs for turning and acceleration.

\noindent \textbf{BarrierNet design.} The robot is required to avoid a circular obstacle in its path, i.e, the state of the robot should satisfy:
\begin{equation}\label{eqn:robotobs}
(x - x_o)^2 + (y - y_o)^2 \geq R^2,
\end{equation}
where $(x_o, y_o)\in\mathbb{R}^2$ denotes the location of the obstacle, and $R > 0$ is the radius of the obstacle.

\begin{figure}[thpb]
	\centering
	\includegraphics[scale=0.35]{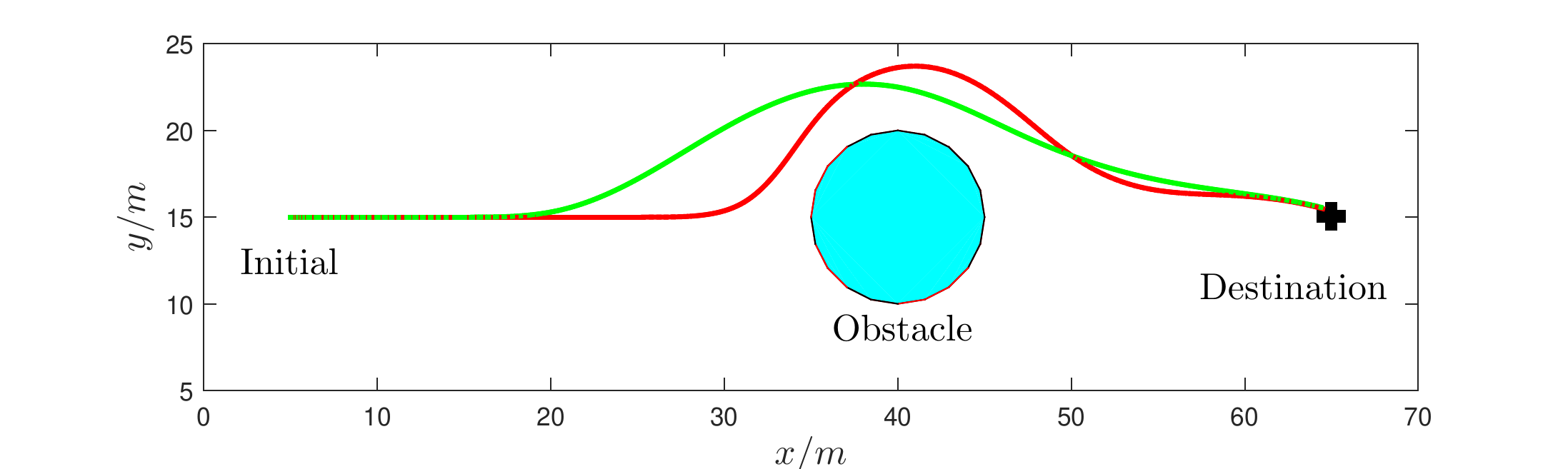}
	\caption{A 2D robot navigation problem. The robot is required to avoid the obstacle in its path. The trajectories (the red and green ones) vary under different definitions of HOCBFs that enforce the safety cosntraint (\ref{eqn:robotobs}).}	
	\label{fig:robot_control}
\end{figure}

The goal is to minimize the control input effort, while subject to the safety constraint (\ref{eqn:robotobs}) as the robot approaches its destination, as shown in Fig. \ref{fig:robot_control}.

The relative degree of the safety constraint (\ref{eqn:robotobs}) is 2 with respect to the dynamics (\ref{eqn:robot}), thus, we use a HOCBF $b(\bm x) = (x - x_o)^2 + (y - y_o)^2 - R^2$ to enforce it. Any control input $\bm u$ should satisfy the HOCBF constraint (\ref{eqn:constraint}) which in this case (choose $\alpha_1, \alpha_2$ in Def. \ref{def:hocbf} as linear functions) is:
\begin{equation}
\begin{aligned}
    -L_gL_fb(\bm x)\bm u\leq L_f^2b(\bm x) + (p_1(\bm z) + p_2(\bm z)) L_fb(\bm x) + (\dot p_1(\bm z) + p_1(\bm z)p_2(\bm z))b(\bm x)
\end{aligned}
\end{equation}
where 

\begin{equation}
    \begin{split}
    & L_gL_fb(\bm x) = [-2(x - x_o)v\sin\theta + 2(y - y_o)v\cos\theta,\quad 2(x - x_o)\cos\theta + 2(y - y_o)\sin\theta]\\
    & L_f^2b(\bm x) = 2v^2\\
    &L_fb(\bm x) = 2(x - x_o) v\cos\theta + 2(y - y_o)v\sin\theta
    \end{split}
\end{equation}

\noindent In the above equations, $\bm z = (\bm x, x_d)$ is the input to the model, 
where $x_d\in\mathbb{R}^2$ is the location of the destination, and $p_1(\bm z), p_2(\bm z)$ are the trainable penalty functions. $\dot p_1(\bm x)$ could be set as 0 due to the discretization solving method of the QP \citep{ames2017control}. 

The cost in the neuron of the BarrierNet is given by:
\begin{equation}
    \min_{\bm u} (u_1 - f_1(\bm z))^2 + (u_2 - f_2(\bm z))^2
\end{equation}
where $f_1(\bm z), f_2(\bm z)$ are references controls provided by the upstream network (the outputs of the FC network).

\begin{figure}[htbp] 
	\centering
	
	\subfigure[Control $u_1$.]{
		\begin{minipage}[t]{0.32\linewidth}
			\centering
			\includegraphics[scale=0.33]{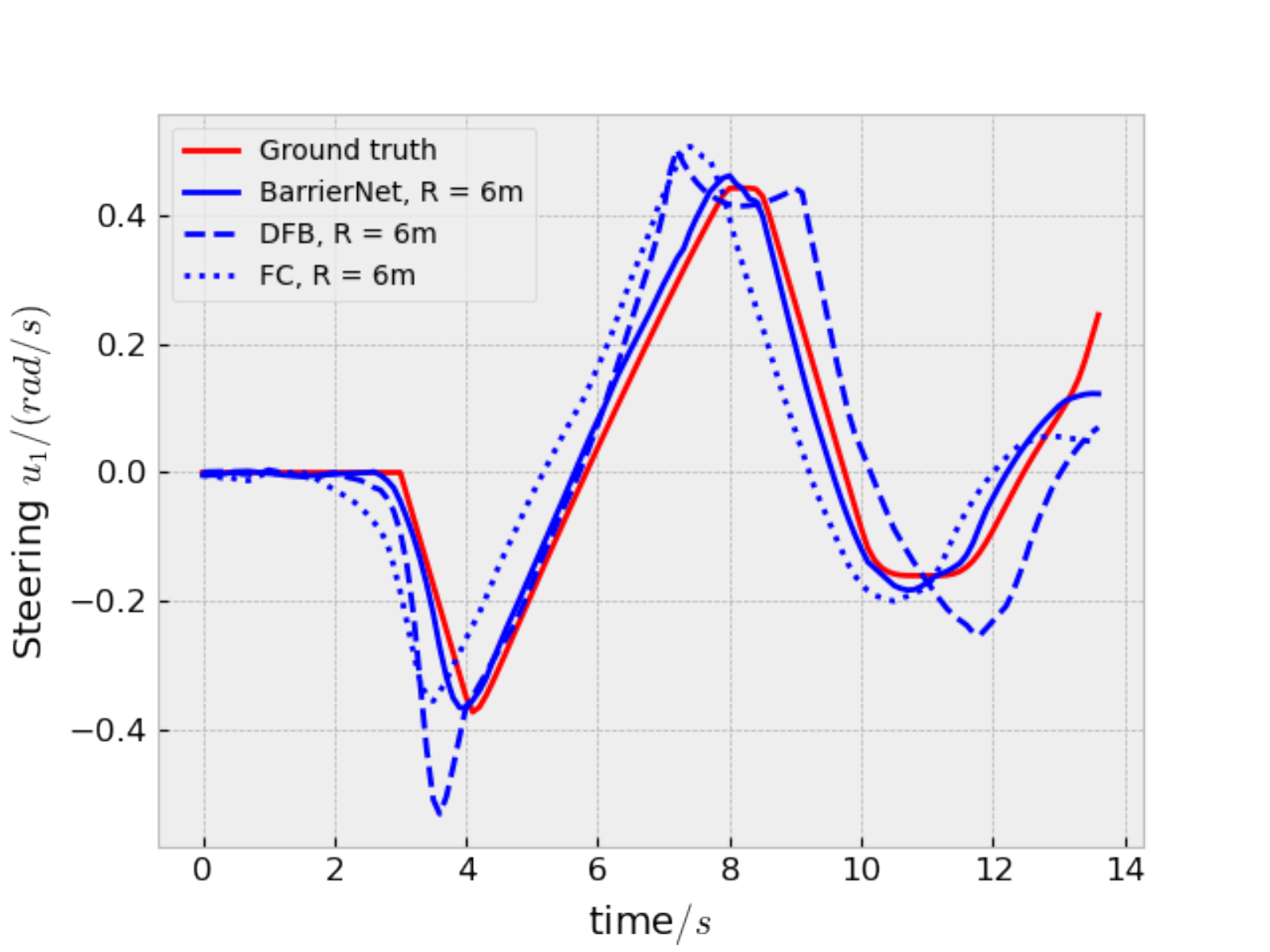} 
			\label{fig:control1_s}%
		\end{minipage}%
	}
	\subfigure[Control $u_2$.]{
		\begin{minipage}[t]{0.32\linewidth}
			\centering
			\includegraphics[scale=0.33]{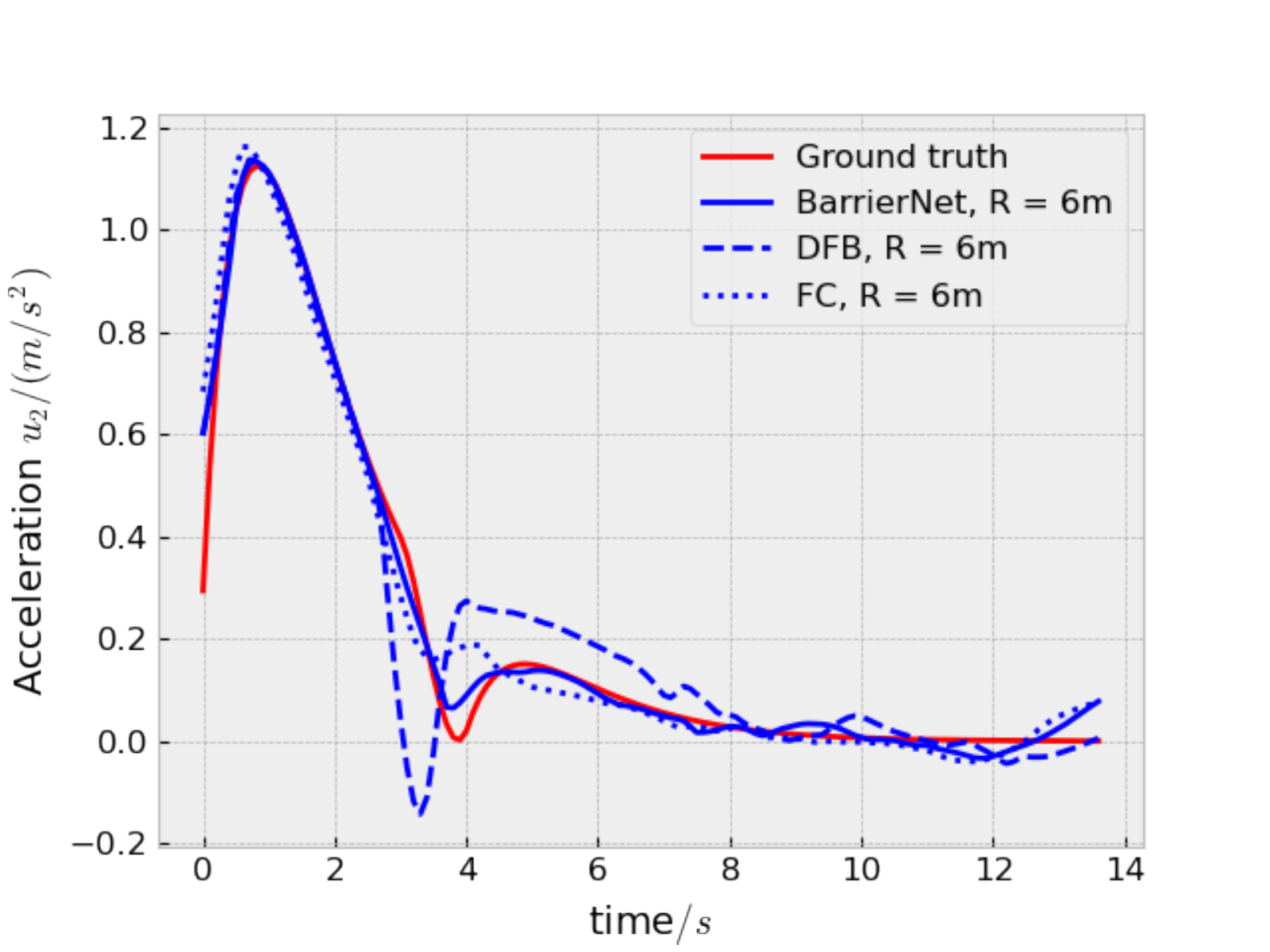} 
			\label{fig:control2_s}%
		\end{minipage}%
	}
	\subfigure[Penalty functions.]{
		\begin{minipage}[t]{0.32\linewidth}
			\centering
			\includegraphics[scale=0.33]{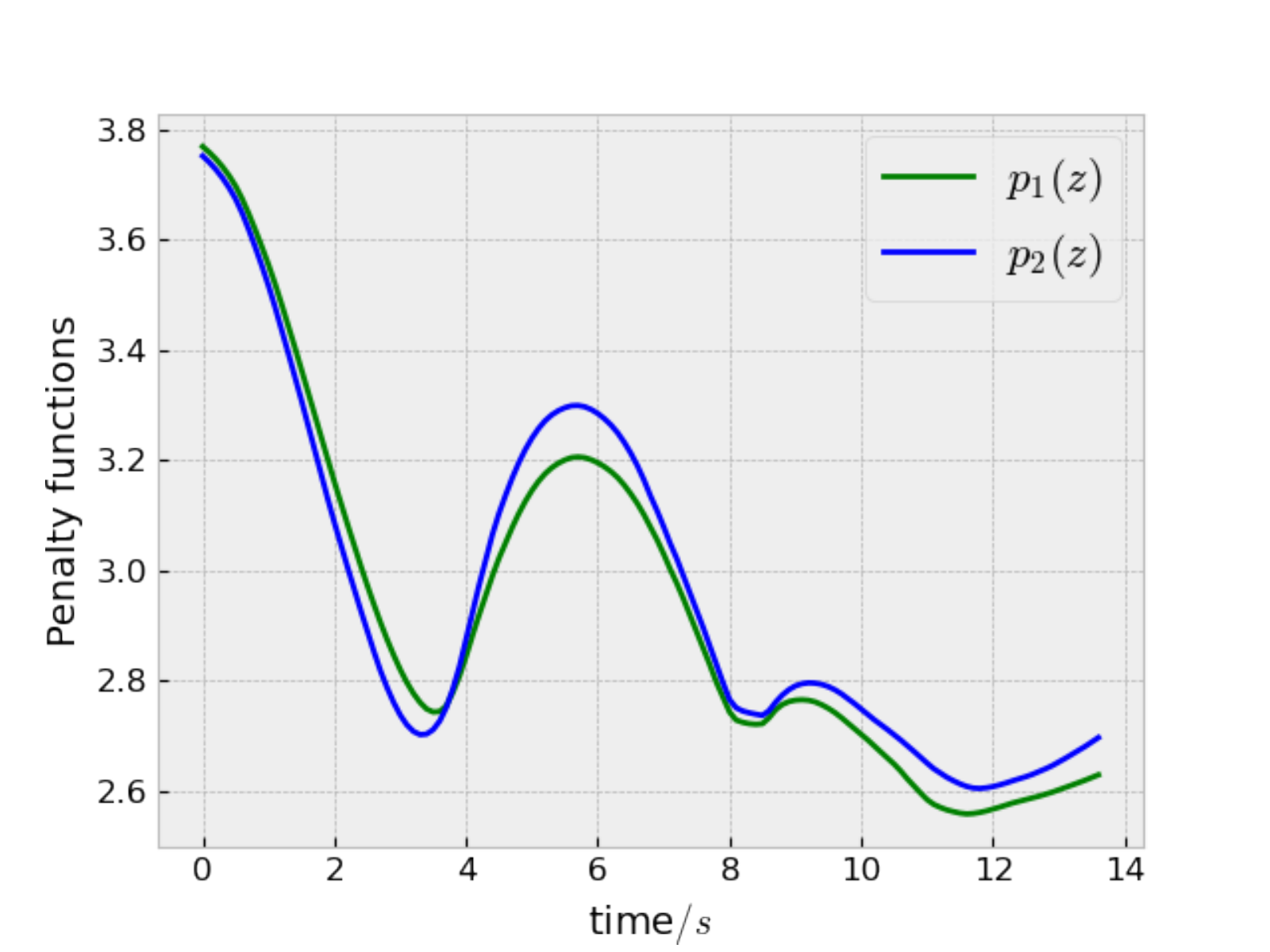} 
			\label{fig:traj_s}%
		\end{minipage}%
	}
	
	\subfigure[Robot trajectories.]{
		\begin{minipage}[t]{0.8\linewidth}
			\centering
			\includegraphics[scale=0.5]{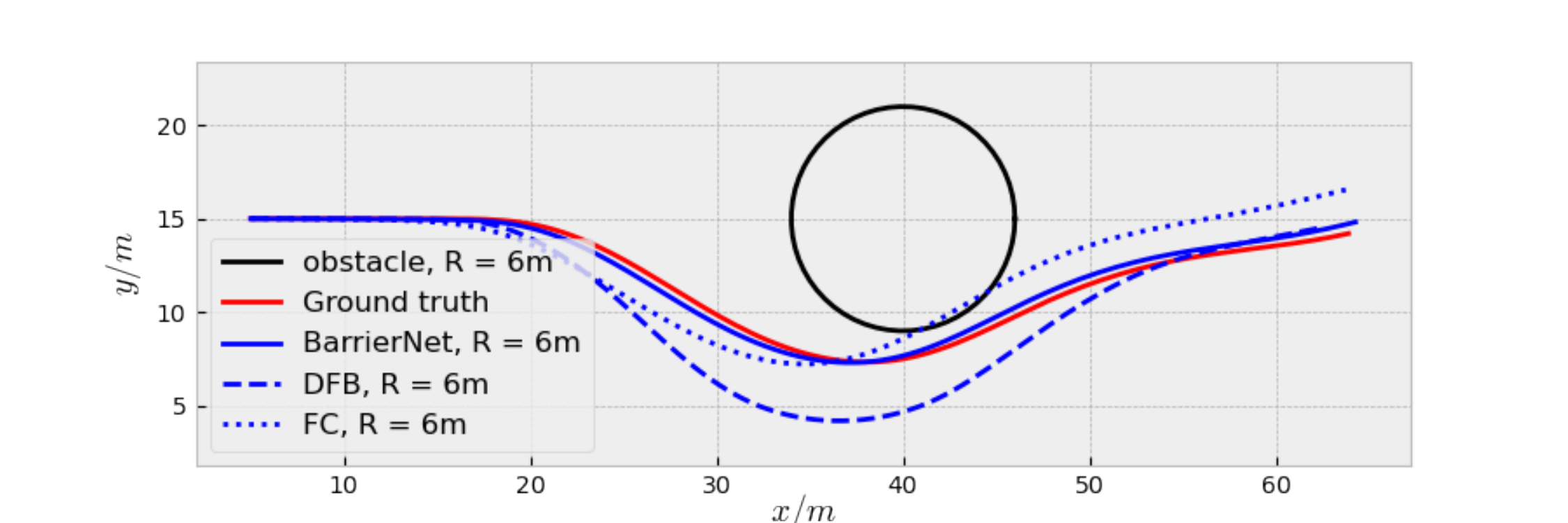} 
			\label{fig:penalty_robot}%
		\end{minipage}%
	}
	
	\centering
	\caption{The controls and trajectories from the FC, DFB and BarrierNet under the training obstacle size $R = 6m$. The results refer to the case that the trained FC/DFB/BarrierNet controller is used to drive a robot to its destination. Safety is guaranteed in both DFB and BarrierNet models, but not in the FC model. The DFB tends to be more conservative such that the trajectories/controls stay away from the ground true as its CBF parameters are not adaptive. The varying penalty functions allow the generation of desired control signals and trajectories (given by training labels), and demonstrate the adaptivity of the BarrierNet with safety guarantees.}
	\label{fig:control_traj}
\end{figure}

\textbf{Results and dicussion.} The training data is obtained by solving the CBF controller introduced in \citep{Xiao2021TAC2}, and we generate 100 trajectories of different destinations as the training data set. We compare the FC model, the deep foward-backward model (referred as DFB) \citep{pereira2020} that is equivalent to take the CBF-based QP as a safety filter, and our proposed BarrierNet. The training and testing results are shown in Figs. \ref{fig:control_traj}a-d. All the models are trained for obstacle size $R = 6m$. The controls from the BarrierNet can stay very close to the ground truth, while there are jumps for controls from the DFB when the robot gets close to the obstacle, which shows the conservativeness of the DFB, as shown by the blue solid (BarrierNet) and blue dashed (DFB) curves in Figs. \ref{fig:control_traj}a and \ref{fig:control_traj}b. The robot trajectory (dashed blue) from the DFB stays far away from ground truth in Fig. \ref{fig:control_traj}d, and this again shows its conservativeness. The robot from the FC will collide with the obstacle as there is no safety guarantee, as the dotted-blue line shown in Fig. \ref{fig:control_traj}d.

\begin{figure}[htbp] 
	\centering
	
	\subfigure[Control $u_1$.]{
		\begin{minipage}[t]{0.45\linewidth}
			\centering
			\includegraphics[scale=0.5]{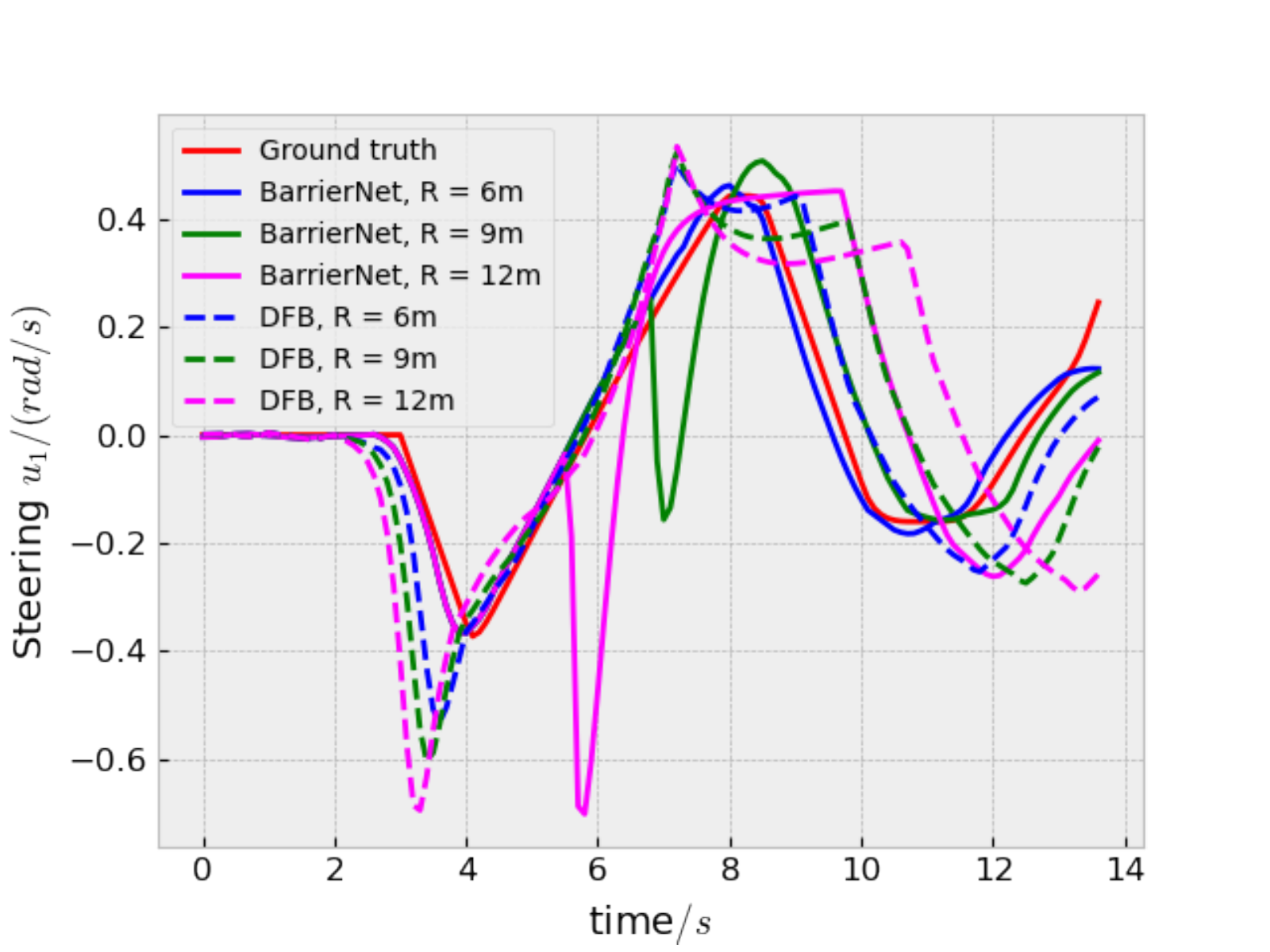} 
			\label{fig:control1_robot}%
		\end{minipage}%
	}
	\subfigure[Control $u_2$.]{
		\begin{minipage}[t]{0.45\linewidth}
			\centering
			\includegraphics[scale=0.5]{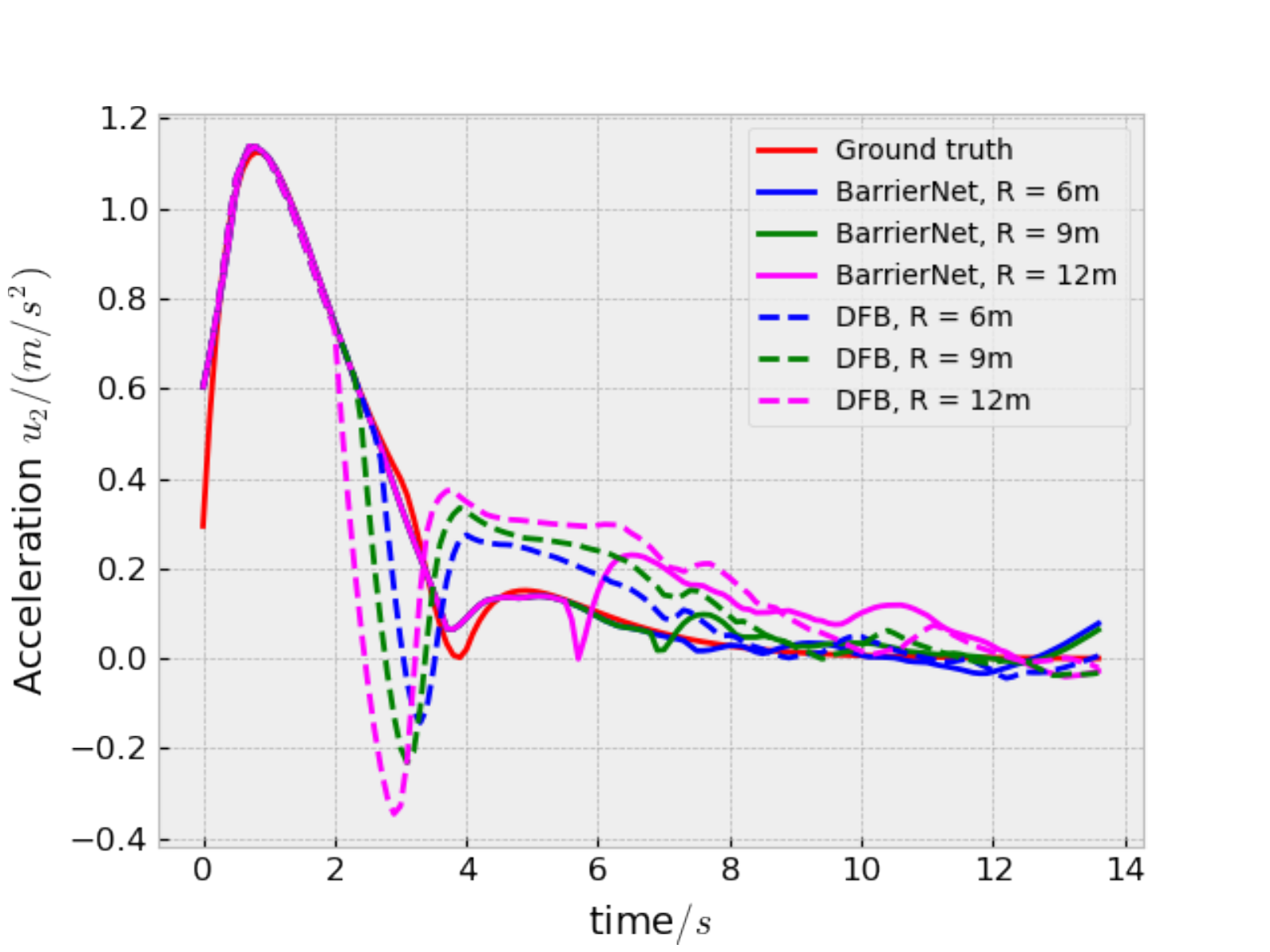} 
			\label{fig:control2_robot}%
		\end{minipage}%
	}
	
	\centering
	\caption{The controls from the BarrierNet and DFB under different obstacle sizes. The BarrierNet and DFB are trained under the obstacle size $R = 6m$. The results refer to the case that the trained BarrierNet/DFB controller is used to drive a robot to its destination. When we increase the obstacle size during implementation, the outputs (controls of the robot) of the BarrierNet and the DFB will adjust accordingly in order to guarantee safety, as shown by the blue and cyan curves. However, the BarrierNet tends to be less conservative for unseen situations.}
	\label{fig:control_mul}
\end{figure}

\begin{figure}[htbp] 
	\centering
	
	\subfigure[The HOCBF $b(\bm x)$ profiles under different obstacle sizes.]{
		\begin{minipage}[t]{0.45\linewidth}
			\centering
			\includegraphics[scale=0.5]{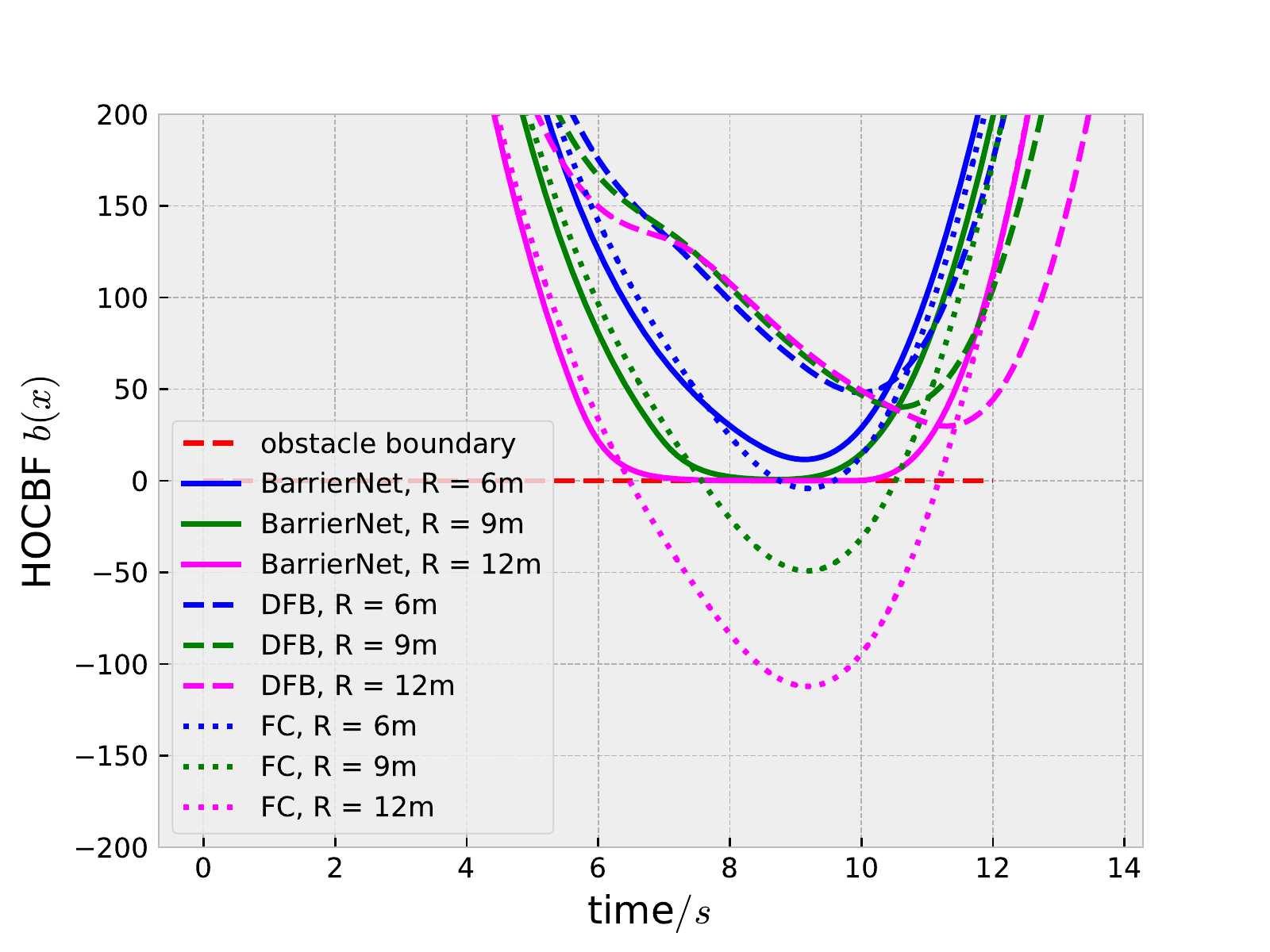} 
		\label{fig:safety_robot}%
		\end{minipage}%
	}
	\subfigure[The robot trajectories under different obstacle sizes.]{
		\begin{minipage}[t]{0.45\linewidth}
			\centering
			\includegraphics[scale=0.5]{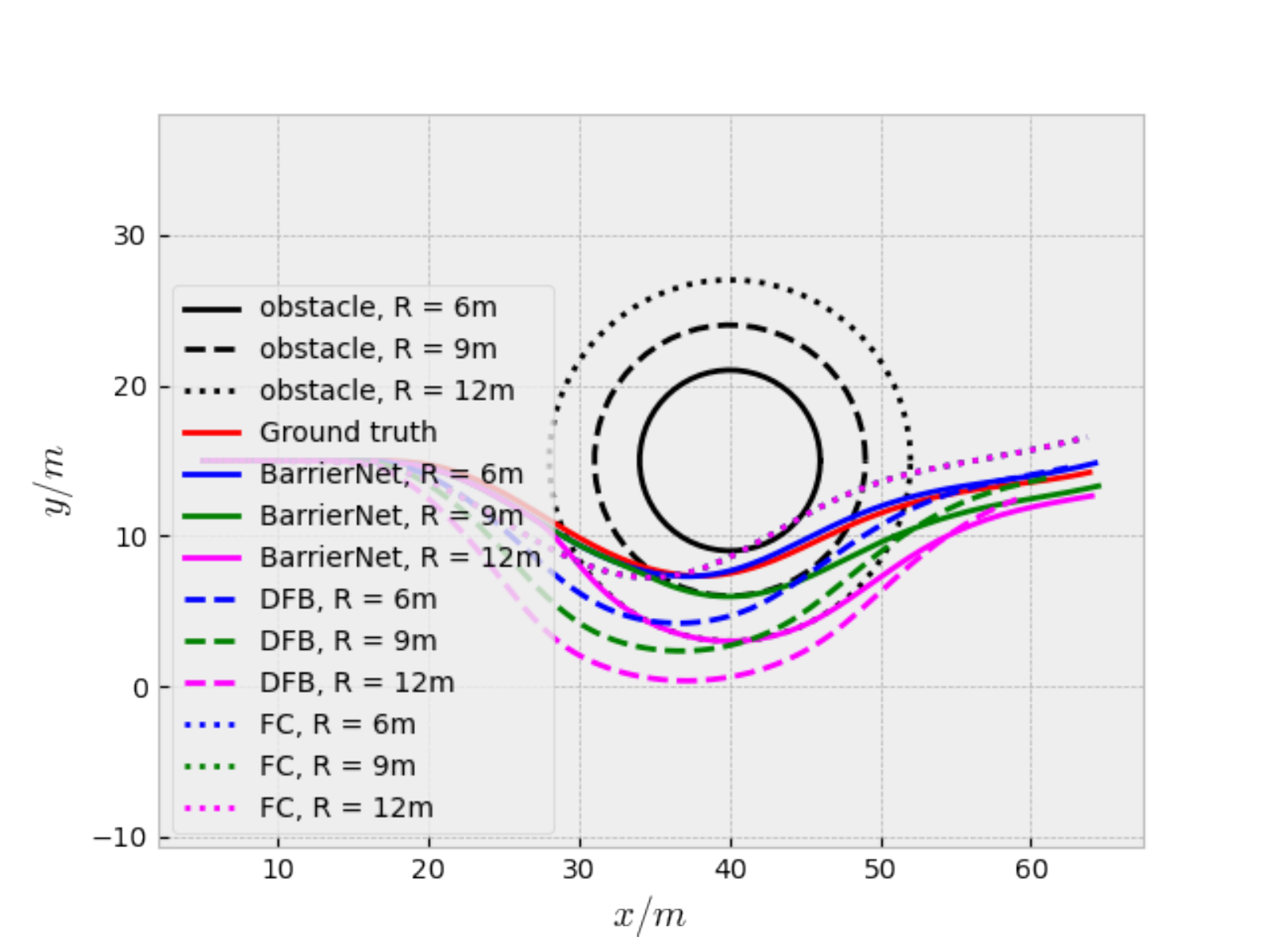} 
			\label{fig:traj}%
		\end{minipage}%
	}
	
	\centering
	\caption{Safety metrics for the BarrierNet, the DFB and the FC network. The BarrierNet, the DFB and the FC network are trained under the obstacle size $R = 6m$. $b(\bm x) \geq 0$ implies safety guarantee. The trajectories under the FC controller coincide as the FC cannot adapt to the size change of the obstacle. }
	\label{fig:safety_mul}
\end{figure}

 When we increase the obstacle size during implementation (i.e., the trained BarrierNet/DFB/FC controller is used to drive a robot to its destination), the controls $u_1, u_2$ from the BarrierNet and DFB deviate from the ground true, as shown in Figs. \ref{fig:control_mul}a and \ref{fig:control_mul}b. This is due to the fact that the BarrierNet and DFB will always ensure safety first. Therefore, safety is always guaranteed in the BarrierNet and DFB, as the solid and dashed curves shown in Fig. \ref{fig:safety_mul}a. Both the BarrierNet and DFB show some adaptivity to the size change of the obstacle. While the FC controller cannot be adaptive to the size change of the obstacle. Thus, the safety constraint (\ref{eqn:robotobs}) will be violated, as shown by the dotted curves in Fig. \ref{fig:safety_mul}a. 

The difference between the DFB and the proposed BarrierNet is in the performance. In Fig. \ref{fig:safety_mul}b, we show all the trajectories from the BarrierNet, DFB and FC controllers under different obstacle sizes. Collisions are avoided under the BarrierNet and DFB controllers, as shown by all the solid and dashed trajectories and the corresponding obstacle boundaries in Fig. \ref{fig:safety_mul}b. However, as shown in Fig. \ref{fig:safety_mul}b,  the trajectories from the BarrierNet (solid) can stay closer to the ground true (red-solid) than the ones from the DFB (dashed) when $R = 6m$ (and other $R$ values). This is due to the fact that the CBFs in the DFB may not be properly defined such that the CBF constraint is active too early when the robot gets close to the obstacle. It is important to note that the robot does not have to stay close to the obstacle boundary under the BarrierNet controller, and this totally depends on the ground truth.  The definitions of CBFs in the proposed BarrierNet depend on the environment (network input), and thus, they are adaptive, and are without conservativeness.

The profiles of the penalty functions $p_1(\bm z), p_2(\bm z)$ in the BarrierNet are shown in Fig. \ref{fig:control_traj}c. The values of the penalty functions vary when the robot approaches the obstacle and gets to its destination, and it shows the adaptivity of the BarrierNet in the sense that with the varying penalty functions, a BarrierNet can produce desired control signals given by labels (ground truth). This is due to the fact the varying penalty functions soften the HOCBF constraint without loosing safety guarantees.


\subsection{3D Robot Navigation}
\noindent\textbf{Experiment setup.} We consider a robot navigation problem with obstacle avoidance in 3D space. In this case, we consider complicated superquadratic safety constraints. The robot navigates according to the double integrator dynamics. The state of the robot is $\bm x = (p_x, v_x, p_y, v_y, p_z, v_z) \in\mathbb{R}^6$, in which the components denote the position and speed along $x, y, z$ axes, respectively. The three control inputs $u_1, u_2, u_3$ are the acceleration along $x, y, z$ axes, respectively.

\noindent \textbf{BarrierNet design.} The robot is required to avoid a superquadratic obstacle in its path, i.e, the state of the robot should satisfy:
\begin{equation}\label{eqn:robotobs3D}
(p_x - x_o)^4 + (p_y - y_o)^4 + (p_z - z_o)^4 \geq R^4,
\end{equation}
where $(x_o, y_o, z_o)\in\mathbb{R}^3$ denotes the location of the obstacle, and $R > 0$ is the half-length of the superquadratic obstacle.

\begin{figure}[thpb]
	\centering
	\includegraphics[scale=0.35]{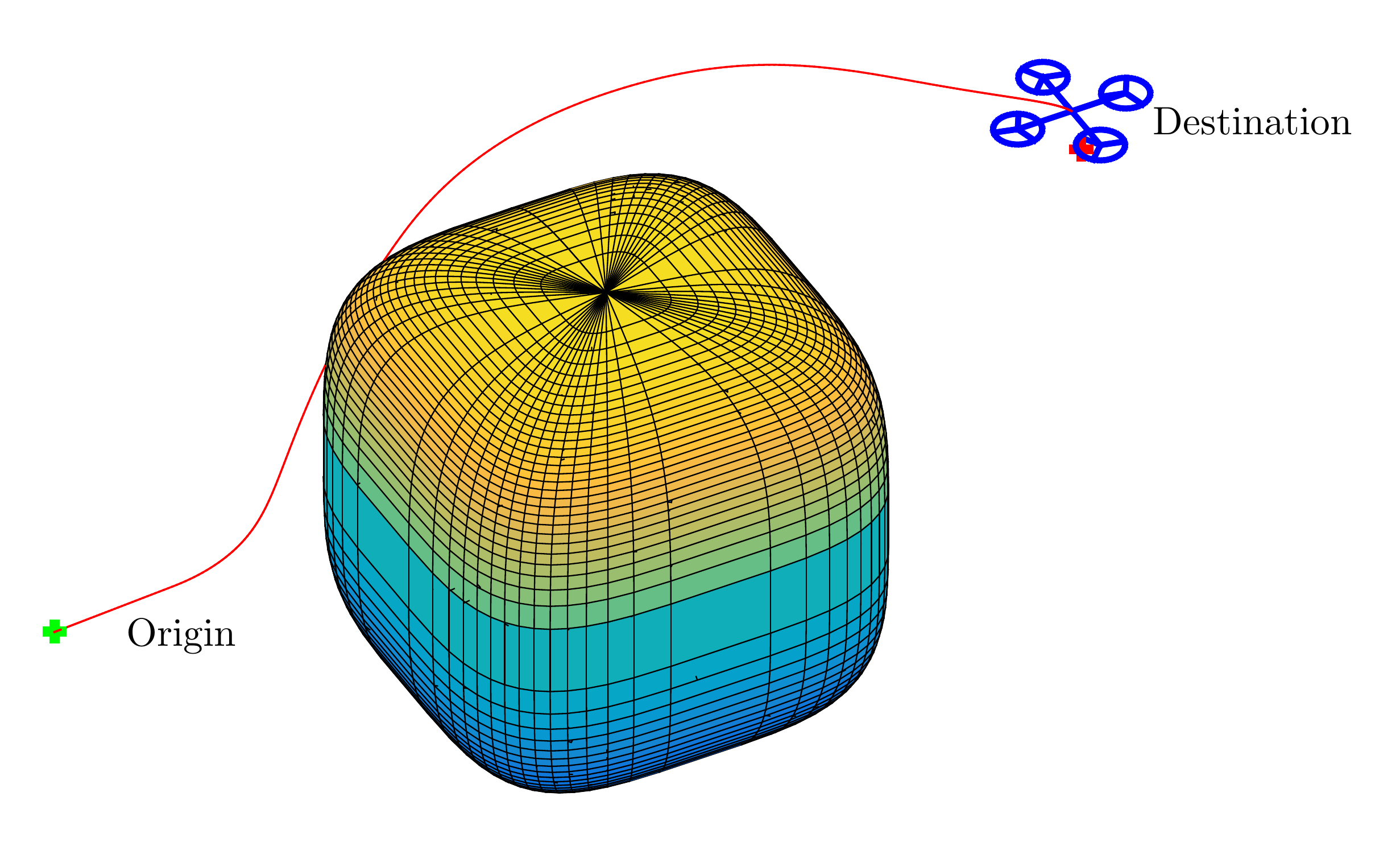}
	\caption{A 3D robot navigation problem. The robot is required to avoid the obstacle in its path. }	
	\label{fig:robot_control3D}
\end{figure}

The goal is to minimize the control input effort, while subject to the safety constraint (\ref{eqn:robotobs3D}) as the robot approaches its destination, as shown in Fig. \ref{fig:robot_control3D}.

The relative degree of the safety constraint (\ref{eqn:robotobs3D}) is 2 with respect to the dynamics, thus, we use a HOCBF $b(\bm x) = (p_x - x_o)^4 + (p_y - y_o)^4 + (p_z - z_o)^4 - R^4$ to enforce it. Any control input $\bm u$ should satisfy the HOCBF constraint (\ref{eqn:constraint}) which in this case (choose $\alpha_1, \alpha_2$ in Def. \ref{def:hocbf} as linear functions) is:
\begin{equation}
\begin{aligned}
    -L_gL_fb(\bm x)\bm u\leq L_f^2b(\bm x) + (p_1(\bm z) + p_2(\bm z)) L_fb(\bm x) + (\dot p_1(\bm z) + p_1(\bm z)p_2(\bm z))b(\bm x)
\end{aligned}
\end{equation}
where 

\begin{equation}
    \begin{split}
    & L_gL_fb(\bm x) = [4(p_x - x_o)^3,\quad 4(p_y - y_o)^3,\quad 4(p_z - z_o)^3]\\
    & L_f^2b(\bm x) = 12(p_x - x_o)^2v_x^2 + 12(p_y - y_o)^2v_y^2 + 12(p_z - x_o)^2v_z^2\\
    &L_fb(\bm x) = 4(p_x - x_o)^3v_x + 4(p_y - y_o)^3v_y + 4(p_z - z_o)^3v_z
    \end{split}
\end{equation}

\noindent In the above equations, $\bm z = \bm x$ is the input to the model, 
 $p_1(\bm z), p_2(\bm z)$ are the trainable penalty functions. $\dot p_1(\bm x)$ is also set as 0 as in the 2D navigation case. 

The cost in the neuron of the BarrierNet is given by:
\begin{equation}
    \min_{\bm u} (u_1 - f_1(\bm z))^2 + (u_2 - f_2(\bm z))^2 + (u_3 - f_3(\bm z))^2
\end{equation}
where $f_1(\bm z), f_2(\bm z), f_3(\bm z)$ are references controls provided by the upstream network (the outputs of the FC network).

\begin{figure}[htbp] 
	\centering
	
	\subfigure[Control $u_1$.]{
		\begin{minipage}[t]{0.22\linewidth}
			\centering
			\includegraphics[scale=0.22]{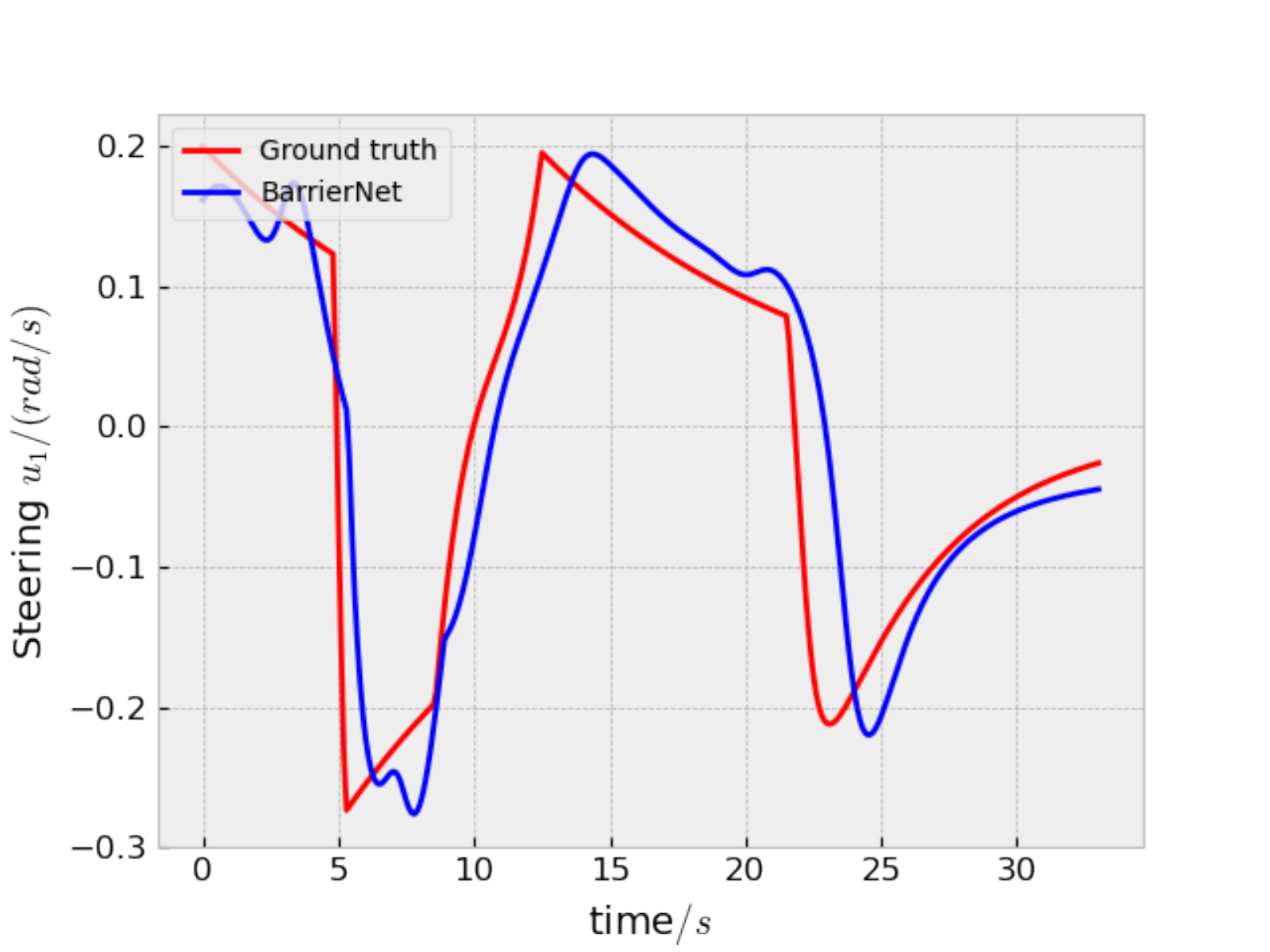} 
		\end{minipage}%
	}
	\subfigure[Control $u_2$.]{
		\begin{minipage}[t]{0.22\linewidth}
			\centering
			\includegraphics[scale=0.22]{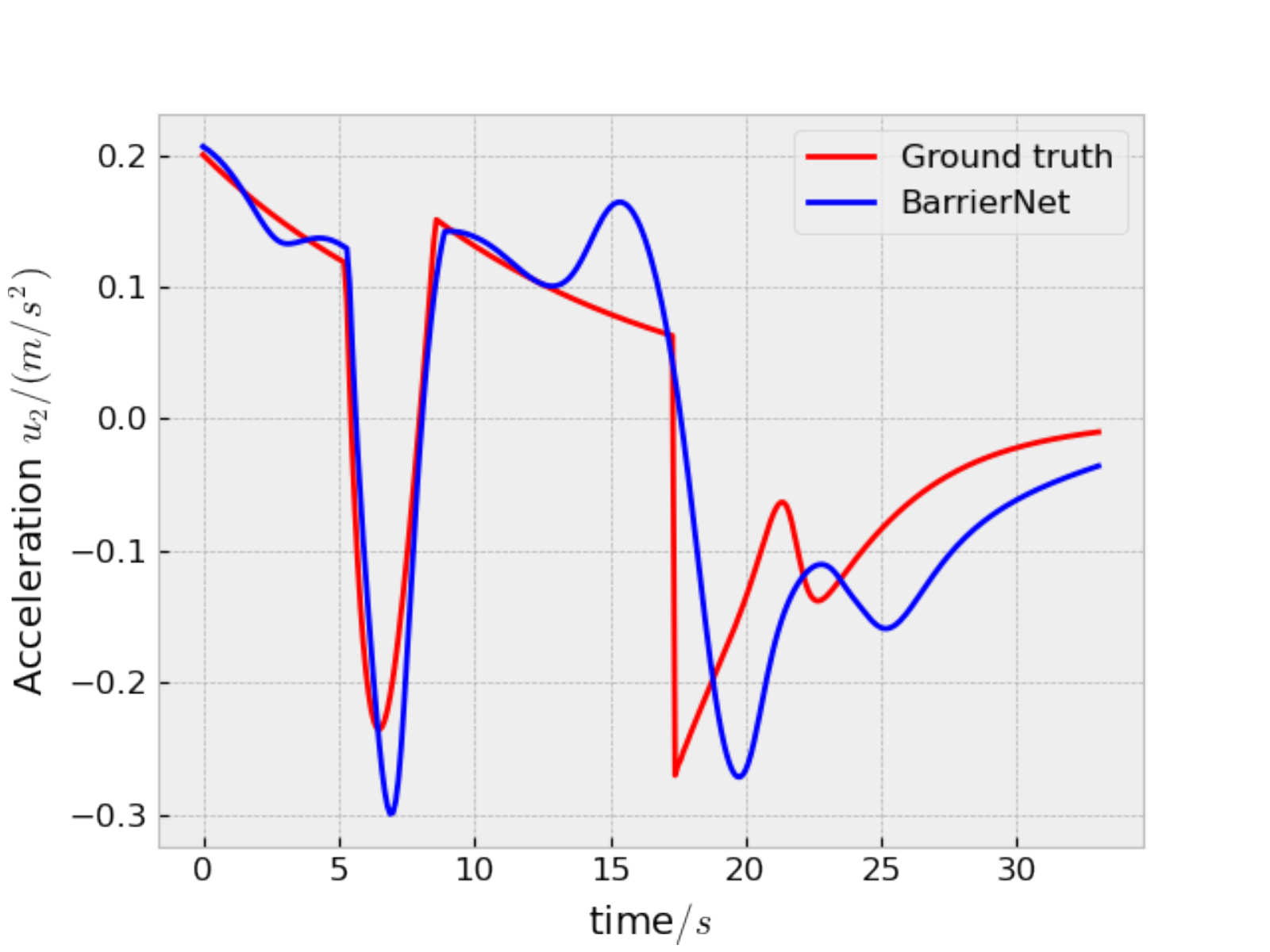} 
		\end{minipage}%
	}
	\subfigure[Control $u_3$.]{
		\begin{minipage}[t]{0.22\linewidth}
			\centering
			\includegraphics[scale=0.22]{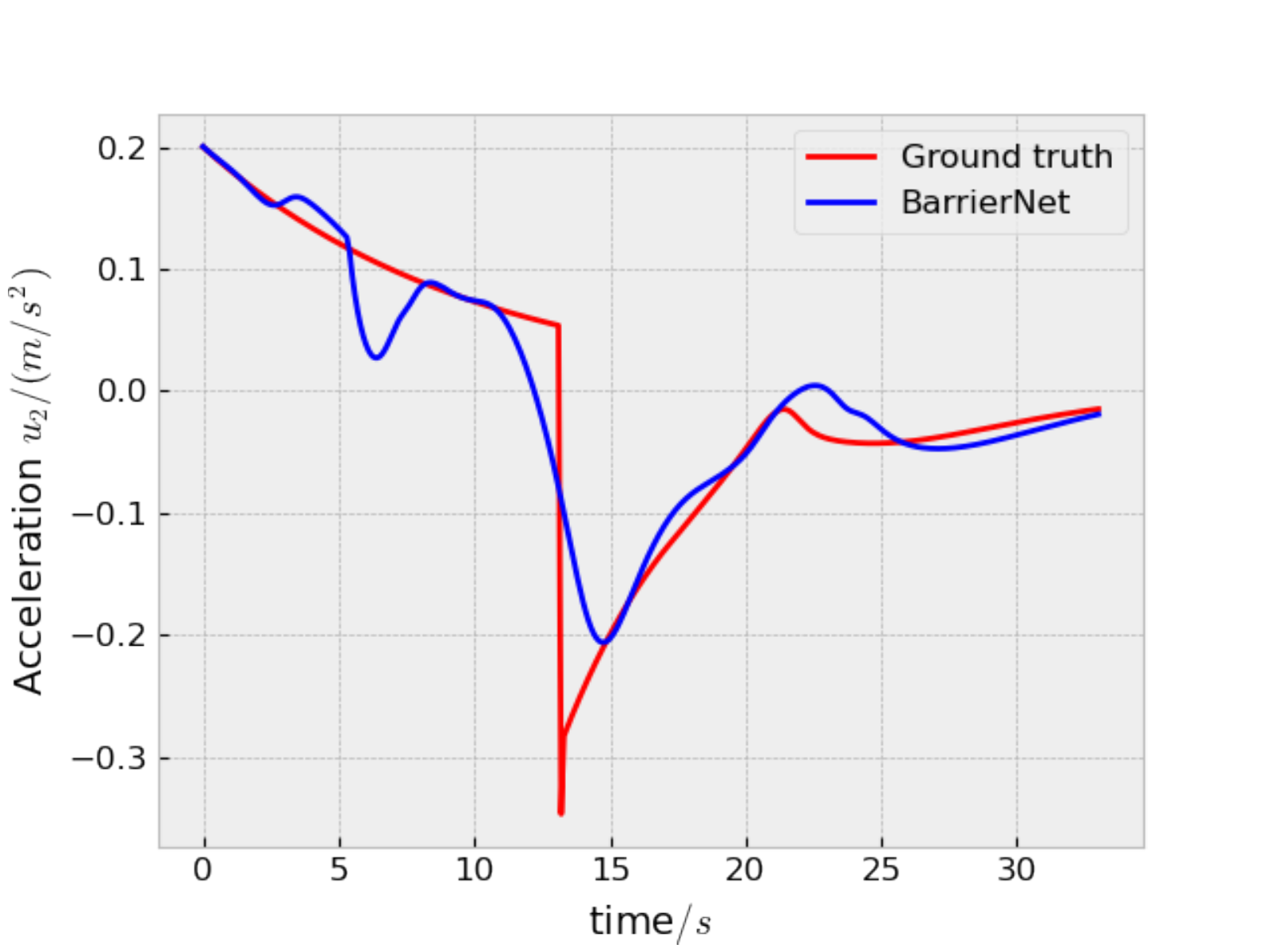} 
		\end{minipage}%
	}
	\subfigure[Penalty functions.]{
		\begin{minipage}[t]{0.22\linewidth}
			\centering
			\includegraphics[scale=0.22]{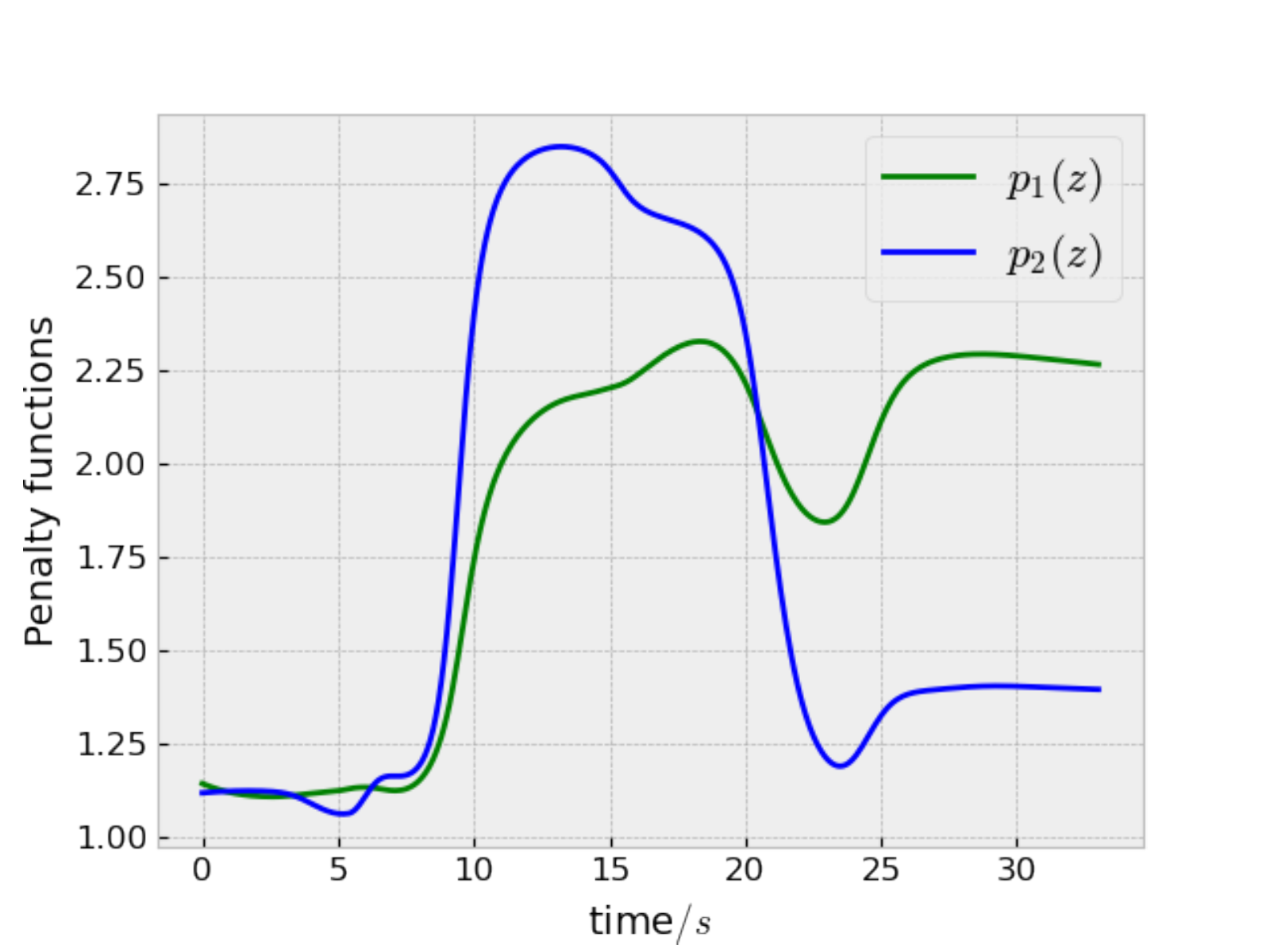} 
		\end{minipage}%
	}

	\centering
	\caption{The controls and penalty functions from the and BarrierNet. The results refer to the case that the trained BarrierNet controller is used to drive a robot to its destination.  The varying penalty functions allow the generation of desired control signals and trajectories (given by training labels), and demonstrate the adaptivity of the BarrierNet with safety guarantees.}
	\label{fig:control3D}
\end{figure}

\begin{figure}[htbp] 
	\centering
	
	\subfigure[The HOCBF $b(\bm x)$ profiles.]{
		\begin{minipage}[t]{0.48\linewidth}
			\centering
			\includegraphics[scale=0.55]{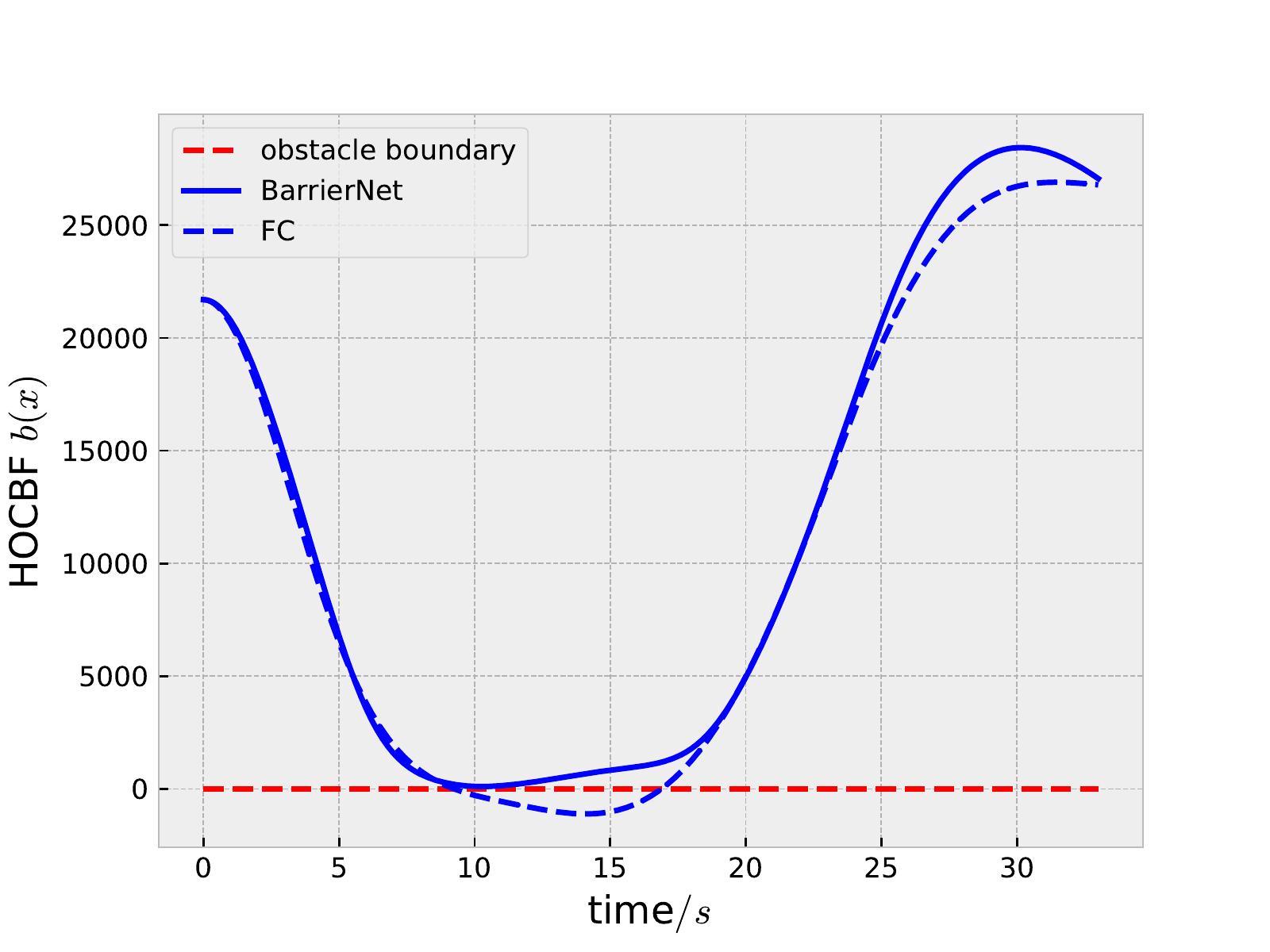} 
		\end{minipage}%
	}
	\subfigure[The robot trajectories.]{
		\begin{minipage}[t]{0.48\linewidth}
			\centering
			\includegraphics[scale=0.55]{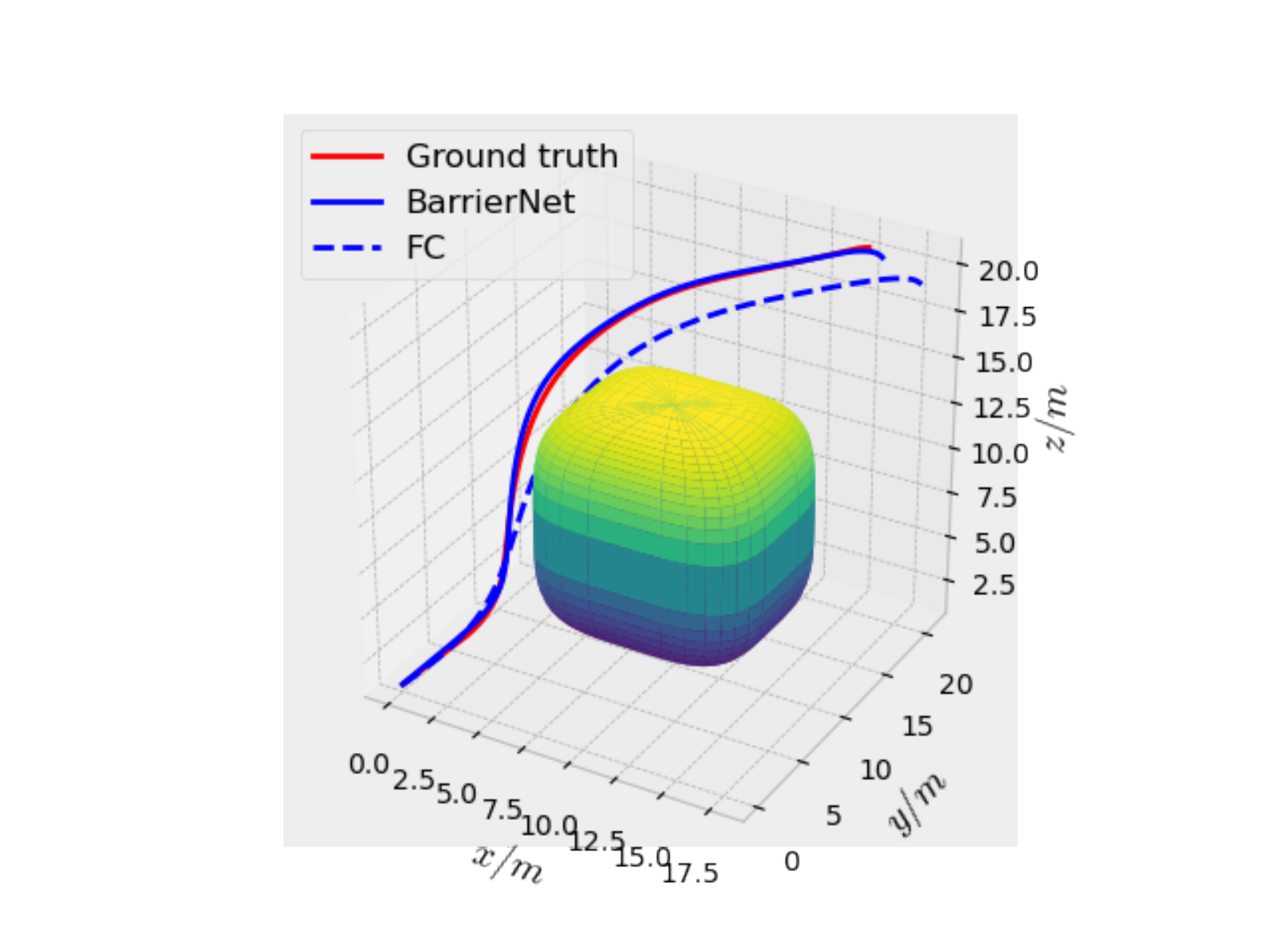} 
		\end{minipage}%
	}

	\centering
	\caption{The HOCBFs and trajectories from the FC and BarrierNet. The results refer to the case that the trained FC/BarrierNet controller is used to drive a robot to its destination. Safety is guaranteed in the BarrierNet model, but not in the FC model. }
	\label{fig:traj3D}
\end{figure}

\textbf{Results and dicussion.}
The training data is obtained by solving the CBF controller introduced in \citep{Xiao2021TAC2}. We compare the FC model with our proposed BarrierNet. The training and testing results are shown in Figs. \ref{fig:control3D} and \ref{fig:traj3D}. The controls from the BarrierNet have some errors with repsect to the ground truth, and this is due to the complicated safety constraint (\ref{eqn:robotobs3D}). We can improve the  tracking accuracy with deeper BarrierNet models (not the focus of this paper). Nevertheless, the implementation trajectory under the BarrierNet controller is close to the ground truth, as shown in Fig.\ref{fig:traj3D}b.

The robot is guaranteed to be collision-free from the obstacle under the BarrierNet controller, as the solid-blue line shown in Fig.\ref{fig:traj3D}b. While the robot from the FC may collide with the obstacle as there is no safety guarantee, as the dotted-blue line shown in Fig. Fig.\ref{fig:traj3D}b.  The barrier function in  Fig.\ref{fig:traj3D}a also demonstrates the safety guarantees of the BarrierNet, but not in the FC model. The profiles of the penalty functions $p_1(\bm z), p_2(\bm z)$ in the BarrierNet are shown in Fig. \ref{fig:control3D}d. The values of the penalty function variations demonstrate the adaptivity of the BarrierNet in the sense that with the varying penalty functions, a BarrierNet can produce desired control signals given by labels (ground truth).

\section{Conclusion}
\label{sec:conc}
In this work, we proposed BarrierNet - a differentiable HOCBF layer that is trainable and guarantees safety with respect to the user defined safe sets. BarrierNet can be integrated with any upstream neural network controller to provide a safety layer. In our experiments, we show that the proposed BarrierNet can guarantee safety while addressing the conservativeness that control barrier functions induce. A potential future avenue of research emerging from this work will be to simultaneously learn the system dynamics and unsafe sets with BarrierNets. This can be enabled using the expressive class of continuous-time neural network models \citep{chen2018neural,lechner2020neural,hasani2021liquid,vorbach2021causal}.

\acks{This research was sponsored by the United States Air Force Research Laboratory and the United States Air Force Artificial Intelligence Accelerator and was accomplished under Cooperative Agreement Number FA8750-19-2-1000. The views and conclusions contained in this document are those of the authors and should not be interpreted as representing the official policies, either expressed or implied, of the United States Air Force or the U.S. Government. The U.S. Government is authorized to reproduce and distribute reprints for Government purposes notwithstanding any copyright notation herein. This work was further supported by The Boeing Company, and the Office of Naval Research (ONR) Grant N00014-18-1-2830. We are grateful to the members of the Capgemini research team for discussing the importance of safety and stability of machine learning.}

\bibliography{MCBF}

\end{document}